\documentclass[11pt]{article}
\usepackage{booktabs}
\usepackage{xspace}

\newcommand{\eg}{{\emph{e.g.}},\xspace}

\usepackage{colortbl}
\usepackage{arydshln}
\usepackage[markup=default]{changes}
\usepackage[final]{acl}
\usepackage{multirow}
\usepackage{times}
\usepackage{latexsym}
 \usepackage[table,xcdraw]{xcolor}
\usepackage[T1]{fontenc}
\usepackage{hyperref} % 导入 hyperref 宏包
\hypersetup{
colorlinks=true, % 链接以颜色显示
urlcolor=blue % 设置超链接颜色为蓝色
}

\usepackage{pifont}
\usepackage{amssymb}
\usepackage[utf8]{inputenc}

\usepackage{microtype}

\usepackage{inconsolata}

\usepackage{graphicx}

\title{SMART: Evaluating LLMs' Mathematical Reasoning via a Human Cognitive Process-Inspired Benchmark}

 \author{Yujie Hou, Mei Wang, Yaoyao Zhong, Ting Zhang, Xuetao Ma, Hua Huang\thanks{Corresponding author} \\
         School of Artificial Intelligence, Beijing Normal University \\  Beijing Key Laboratory of Artificial Intelligence for Education \\ Engineering Research Center of Intelligent Technology and Educational Application, Ministry of Education\\
        \texttt{\{houyujie, maxuetao\}@mail.bnu.edu.cn, \{wangmei1, zhongyy, tingzhang, huahuang\}@bnu.edu.cn}}

\begin{document}
\maketitle
\begin{abstract}
Large Language Models (LLMs) have achieved remarkable performance across a wide range of mathematical benchmarks. However, concerns remain as to whether these successes reflect genuine reasoning or superficial pattern recognition. Existing evaluation methods, which typically focus either on the final answer or on the intermediate reasoning steps, reduce mathematical reasoning to a shallow input–output mapping, overlooking its inherently multi-stage and multi-dimensional cognitive nature. Inspired by P\'olya’s problem-solving theory, we propose SMART, a benchmark that decomposes mathematical problem-solving into four cognitive dimensions: \textbf{S}emantic Understanding, \textbf{M}athematical Reasoning, \textbf{A}rithmetic Computation, and \textbf{R}eflection \& Refinemen\textbf{t}, and introduces dimension-specific tasks to measure the corresponding cognitive processes of LLMs. We apply SMART to 22 state-of-the-art open- and closed-source LLMs and uncover substantial discrepancies in their capabilities across dimensions. Our findings reveal genuine weaknesses in current models and motivate a new metric, the All-Pass Score, designed to better capture true problem-solving capability. Data is available at \href{https://huggingface.co/datasets/ewdfd/SMART}{https://huggingface.co/datasets/ewdfd/SMART}.
\end{abstract}

\section{Introduction}
Large language models (LLMs)~\cite{achiam2023gpt,wei2022chain} have demonstrated impressive performance and are being increasingly integrated into real-world applications \eg education~\cite{wang2024large}, scientific computing~\cite{ma-etal-2025-problem}, and decision support~\cite{OpenAI2024,guo2025deepseek}. With this widespread adoption, assessing their capability boundaries has become essential. The mathematical reasoning, a key indicator of higher-order cognition, serves as a critical benchmark to evaluate the logical thinking and systematic problem-solving of models.

\begin{figure}[t]
\centering
  \includegraphics[width=0.95\linewidth]{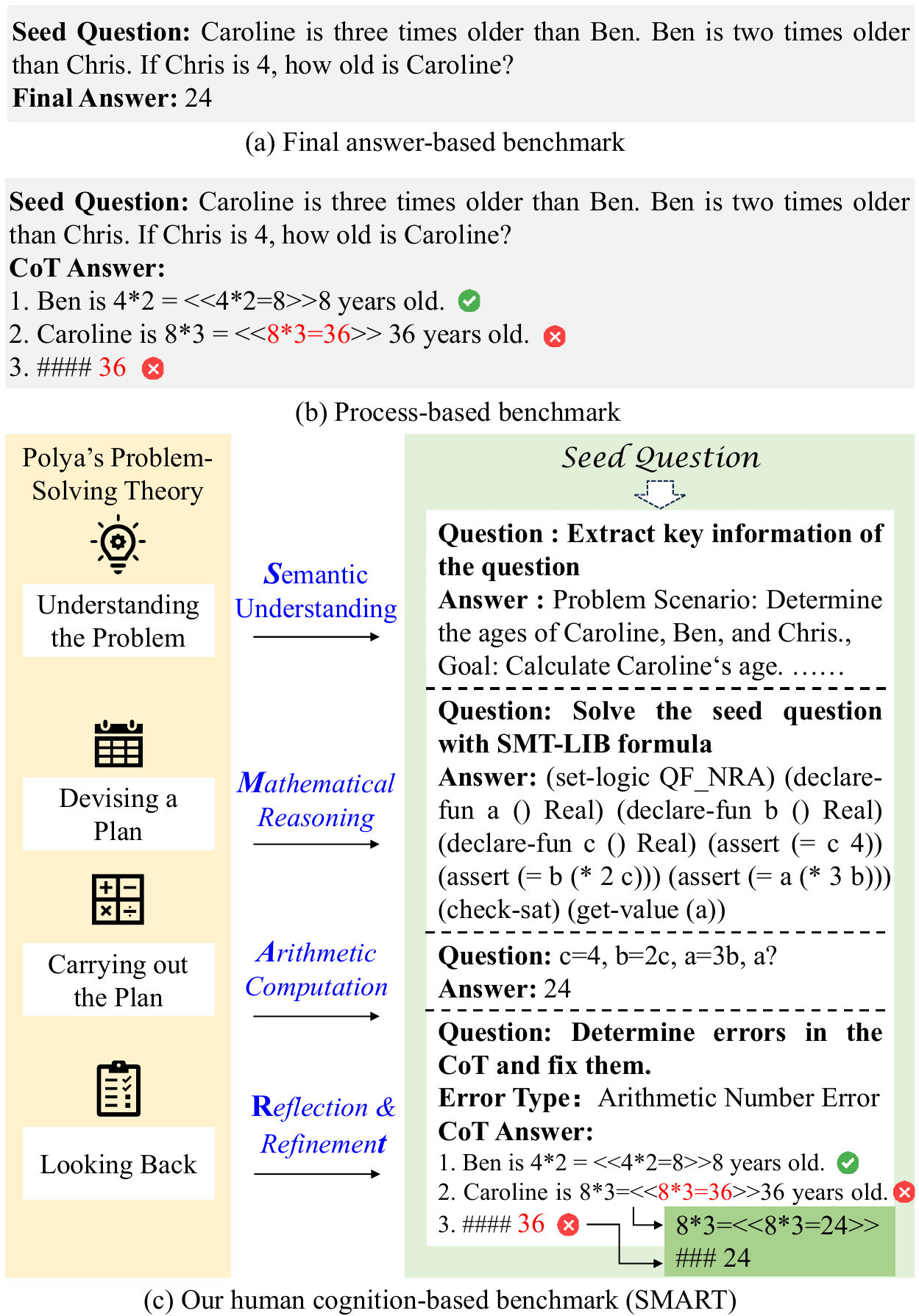}

  \caption{Comparison of evaluation paradigms for LLM mathematical reasoning.
Final-answer-based benchmarks evaluate only the final outcome, process-based benchmarks detect errors in reasoning steps, while SMART builds on P\'olya’s problem-solving theory to evaluate four cognitive dimensions.}
  \label{fig:mot}

\end{figure}

However, existing LLM mathematical benchmarks are misaligned with the human multi-dimensional cognitive process of mathematical problem-solving. P\'olya’s problem-solving theory~\cite{polya2014solve} formalizes this cognitive process into four progressive dimensions: understanding the problem, devising a plan, executing the plan, and looking back on the solution. Unfortunately, mainstream evaluation approaches, such as GSM8K~\cite{cobbe2021training} and MATH~\cite{hendrycks2measuring}, reduce this process to simple end-to-end matching, assessing LLMs solely based on final answer correctness (Fig.\ref{fig:mot}a). While recent benchmarks, MR-GSM8K~\cite{zengmr} and ProcessBench \cite{zheng-etal-2025-processbench}, have begun incorporating step-by-step solution verification, they still fall short of comprehensively evaluating the distinct cognitive stages that underlie mathematical reasoning (Fig.\ref{fig:mot}b). These two approaches fail to capture the subtle cognitive processes at each problem-solving phase, making it impossible to pinpoint where models struggle in the reasoning processes and, therefore limiting guidance for targeted improvements.

To address these limitations, we propose the first benchmark, called SMART, to systematically evaluate the complete cognitive process of LLMs in mathematical reasoning (Fig.\ref{fig:mot}c). Guided by P\'olya's problem-solving theory, SMART systematically decomposes each mathematical problem along the reasoning pipeline into four cognitive dimensions, corresponding to \textbf{S}emantic Understanding (Understanding), \textbf{M}athematical Reasoning (Reasoning), \textbf{A}rithmetic Computation (Arithmetic), and \textbf{R}eflection \& Refinemen\textbf{t} (R\&R). This decomposition enables an independent assessment of LLM capabilities in each cognitive dimension, allowing a fine-grained diagnosis of model performance in distinct problem-solving stages. Moreover, we introduce a new metric, the All-Pass Score, which measures model accuracy only when all four dimension-specific tasks are correctly solved.

Creating a comprehensive, multi-task benchmark at scale presents a fundamental challenge: each problem requires carefully designed sub-questions that target the specific cognitive process, demanding extensive human annotation. To make this approach both scalable and cost-effective, we further introduce an automated generation pipeline that transforms seed problems into four-dimensional assessment tasks, incorporating neuro-symbolic ~\cite{barrett2010smt,de2008z3} and human verification to enable iterative quality validation. Furthermore, these dimension-specific tasks are novel for LLMs and thus contribute to mitigating data contamination. 

We evaluate 22 recently released open- and closed-source LLMs on SMART. Experimental results demonstrate that even the most advanced models perform poorly under the All-Pass Score metric, underscoring the challenging nature of our benchmark. In addition, SMART serves as a diagnostic tool, identifying which cognitive dimensions emerge as the primary bottlenecks in mathematical problem-solving. Furthermore, we find that targeted improvements in specific weak dimensions can lead to substantial overall gains in mathematical capability—for example, increasing the final answer accuracy of Qwen2.5-72B by 11.77\%. Our main contributions are as follows:    
\begin{enumerate}
\item[$\bullet$] To evaluate the true mathematical reasoning capability of LLMs, we propose the SMART benchmark that consists of 10,000 questions across distinct cognitive dimensions, and the new All-Pass Score, enabling comprehensive evaluation of the problem-solving process.

\item[$\bullet$] Of equal importance to the SMART benchmark, we introduce a novel data curation and quality control framework that automates the construction of dimension-specific sub-tasks from seed questions and validates the benchmark via rigorous correctness assessment.

\item[$\bullet$]Based on SMART, we reveal substantial disparities in LLMs’ mathematical capabilities and offer dimension-specific, interpretable diagnostics that pinpoint weaknesses. Targeting the weakest dimension with a reflection-and-refinement prompt boosts Qwen2.5-72B’s final-answer accuracy by 11.77\%.
                     
\end{enumerate}

\begin{figure*}[t]
\centering
  \includegraphics[width=0.95\linewidth]{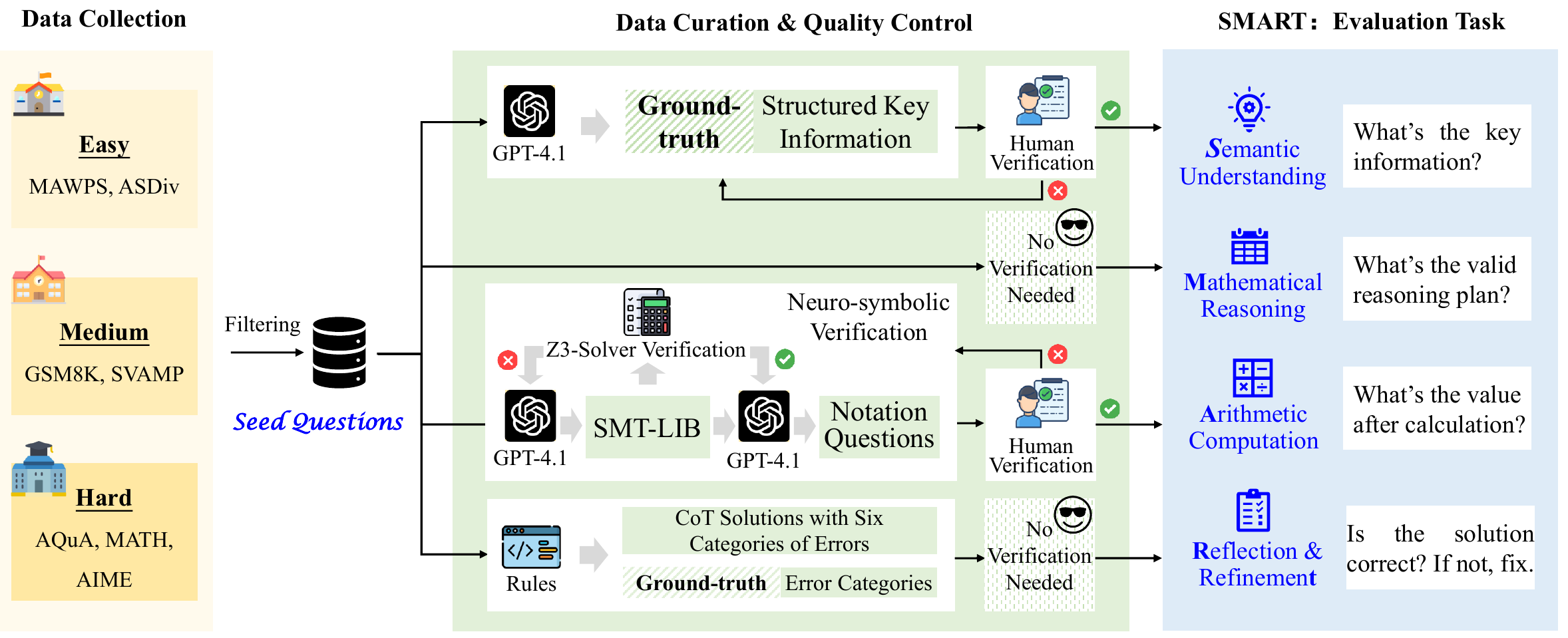}

  \caption{Overview of SMART benchmark construction. First, we collect seed questions from datasets of varying difficulty and filter out those that do not meet our requirements. Second, the seed questions are used to generate dimension-specific tasks along with their corresponding ground-truths. Finally, the generated data are validated through a neuro-symbolic and human verification when needed to ensure data quality. }

  \label{fig:pipline}

\end{figure*}

\section{Related Work}
\textbf{Mathematical Benchmark.} Numerous mathematical benchmarks with varying levels of difficulty have been developed to explore the upper bound of LLMs' mathematical capabilities. These benchmarks range from grade-school-level datasets~\cite{cobbe2021training}, to high-school-level datasets~\cite{hendrycks2measuring}, and extend to expert-level datasets~\cite{glazer2024frontiermath}. Their scope covers a broad range of mathematical domains, including geometry, number theory, and real analysis. However, despite their increasing difficulty, these benchmarks primarily adopt a final answer-based evaluation approach, making it unclear whether LLMs genuinely understand mathematical concepts or simply rely on pattern-matching to produce correct answers~\cite{mirzadehgsm}. To address this, ProcessBench~\cite{zheng-etal-2025-processbench} and PRMBench~\cite{song-etal-2025-prmbench} have been innovatively proposed to enable process-based evaluation by identifying erroneous steps in the model's mathematical reasoning. Nevertheless, these process-based benchmarks still fall short of capturing human thinking, since they do not evaluate the fine-grained cognitive processes across the stages of problem-solving.
      
\noindent\textbf{Dynamic evaluation.} The widespread use of benchmarks increases the risk of data contamination, potentially inflating performance evaluations~\cite{li2024perteval}. Recent studies address these concerns with dynamic evaluation~\cite{zhu2023dyval,zhu2024dynamic} that generate adaptive test data via predefined rules. GSM-Plus~\cite{li2024gsm} and GSM-Symbolic~\cite{mirzadehgsm} similarly generate variants from seed questions. These approaches have shown encouraging progress in mitigating data leakage and improving robustness in evaluations. However, manually annotating newly generated questions is labor-intensive and costly, motivating the need for automated, scalable data generation and verification.

Despite these recent advances in mathematical benchmarks and dynamic evaluation, persistent limitations underscore the need for a benchmark that comprehensively assesses the entire problem-solving process, provides fine-grained and interpretable analyses, and reduces the cost of constructing benchmarks. To address this gap, SMART is designed to systematically evaluate the mathematical reasoning capabilities of LLMs.

\begin{table*}[]
\resizebox{\textwidth}{!}{%
\begin{tabular}{l|cccc}
\toprule[1.5pt]

Task       &   Question (Verification)          & Answer                                                                & Evaluator                          &Ground-truth (Verification)              \\ \hline
Understanding & SQ (\ding{53})             & SKI & LLM-as-a-Judge                                                   & SKI (GPT + Human) \\
Reasoning     & SQ (\ding{53})             & SMT-LIB                                                               & Z3 Solver + Rule-match &Final answer of SQ (\ding{53})                 \\
Arithmetic    & NQ (GPT + Human)  & Answer of NQ                                                          & Rule-match                                                       & Final answer of SQ (\ding{53})                 \\
Reflection    & CoT with Errors (\ding{53}) & Error Categories                                                      & Rule-match                                                       &  Error Categories  (\ding{53})                 \\
Refinement    & CoT with Errors (\ding{53}) & Refined CoT                                                           & Rule-match                                                       & Final answer of SQ (\ding{53})               \\ 
\bottomrule[1.5pt]
\end{tabular}}
           \caption{Overview of question, answer, evaluator, and ground-truth of each dimension task in SMART. SQ means seed question. NQ means notation-based arithmetic question. SKI means structured key information. (\ding{53}) means no verification. }
\label{tab.sum_smart}

\end{table*}

\section{The SMART Benchmark}
\subsection{Overview}
SMART is a fine-grained multi-task benchmark for evaluating the problem-solving capabilities of LLMs. Its four sub-tasks are derived from P\'olya’s problem-solving theory. In \textit{How to Solve It}~\cite{polya1957solve}, P\'olya conceptualized mathematical problem-solving as a four-step cognitive process: (1) Understanding the problem, (2) Devising a plan, (3) Carrying out the plan, and (4) Looking back.
Adopting Pólya’s problem-solving theory can clarify where LLMs succeed or fail by separating cognitive dimensions. Therefore, building on this cognitive framework, SMART evaluates LLMs along four corresponding cognitive dimensions: Semantic Understanding (Understanding), Mathematical Reasoning (Reasoning), Arithmetic Computation (Arithmetic), and Reflection \& Refinement (R\&R). The evaluation settings for the four dimension tasks are summarized in Tab.~\ref{tab.sum_smart}.

\subsection{Evaluation Sub-Tasks}

\noindent \textbf{Understanding.} The Understanding task evaluates a model’s semantic understanding capability by extracting and organizing  key information  from the question. In this task, the input question is a seed question, and the model identifies and categorizes essential components into a predefined template. The template comprises five categories: problem scenario, goal, known and unknown quantities, relationships and constraints, and irrelevant information. 

We adopt this design to evaluate not only the model’s capacity to summarize and highlight salient information but also its depth of comprehension. By requiring the model to distinguish the roles of different elements and their interconnections, the task provides a nuanced measure of problem understanding.

\noindent \textbf{Reasoning.} The Reasoning task evaluates the mathematical reasoning capability by requiring LLMs to produce a symbolic formalization of the solution. Given a seed question, the model is prompted to solve it using symbolic formalization in the SMT-LIB format~\cite{barrett2010smt}. 

This task compels the model to capture the underlying logical structure of the problem and the intricate relationships among its components. With few-shot prompting, LLMs easily learn to produce SMT-LIB–formatted answers.

\noindent \textbf{Arithmetic.} The Arithmetic task evaluates an LLM’s capability to perform arithmetic computation by requiring it to solve notation-based questions containing only numerical values and variables. These notation-based questions  are simplified from the seed questions, expressed purely in terms of numbers and variables, and require the execution of basic arithmetic operations.

We design this task to isolate arithmetic skills from other cognitive demands—such as language comprehension or complex reasoning—thereby providing a focused and precise assessment of a model’s arithmetic capabilities.

\noindent \textbf{R \& R.} The Reflection \& Refinement task evaluates the LLM’s capacity for self-critique. The model is presented with a question and its chain-of-thought (CoT) solution, and is tasked with identifying potential errors in CoT (Reflection). It then revises the errors and produces a refined CoT (Refinement). Importantly, if the model fails to detect all errors during the Reflection stage, it is not allowed to proceed to Refinement.

\subsection{Benchmark Construction}
As shown in Fig.~\ref {fig:pipline}, SMART is constructed in three stages: data collection, data curation, and quality control. Through this deliberately designed pipeline, we can automatically generate four dimension-specific tasks with corresponding ground truths, while requiring significantly less human verification compared to traditional benchmarks that rely heavily on manual annotation.
\subsubsection{Data Collection }
We begin by collecting a diverse set of seed questions from seven widely used mathematical problem datasets spanning three difficulty levels. Easy questions are drawn from MAWPS~\cite{koncel2016mawps} and ASDiv~\cite{miao2020diverse}; medium questions from GSM8K~\cite{cobbe2021training} and SVAMP~\cite{patel2021nlp}; and hard questions from AQuA~\cite{ling2017program}, MATH~\cite{hendrycks2measuring}, and AIME 2024~\cite{aime2024}.

To ensure verifiability and sufficient reasoning complexity, we filter questions that can be formalized in SMT-LIB format (so their solutions can be validated using the Z3 solver) and require at least two reasoning steps, preventing the SMART benchmark from being overly trivial. After filtering, we obtain 2,000 seed questions, which serve as the foundation for constructing the dimension-specific evaluation tasks.

\begin{table*}[]
\resizebox{\textwidth}{!}{%
\renewcommand\arraystretch{1.1}
\begin{tabular}{lcccccccc}
\toprule[1.5pt]
\multicolumn{1}{c|}{\multirow{2}{*}{Model}} &
  Understanding &
  Reasoning &
  Arithmetic &
  \multicolumn{1}{c|}{R\&R} &
  Reflection &
  \multicolumn{1}{c|}{Refinement} &
  Final Answer &
  All-Pass Score \\
\multicolumn{1}{c|}{}                        & LLM@Un & ACC@Re & ACC@Ar & \multicolumn{1}{c|}{ACC@RR} & ACC@R-t & \multicolumn{1}{c|}{ACC@R-m} & ACC@Fi & ACC@All \\ \hline
\multicolumn{9}{c}{Open-source models}                                                                                                                            \\ \hline
\multicolumn{1}{l|}{Phi4-14B}                & 93.41  & 64.12  & 96.45  & \multicolumn{1}{c|}{22.06}  & 26.75   & \multicolumn{1}{c|}{82.34}   & 89.55  &  20.77    \\
\multicolumn{1}{l|}{Gemma3-27B}              & 94.31  & 40.81  & 39.94  & \multicolumn{1}{c|}{24.23}  & 32.55   & \multicolumn{1}{c|}{74.35}   & 54.62  &  11.96       \\
\multicolumn{1}{l|}{GLM4-32B}                & 94.11  & 55.92  & 36.62  & \multicolumn{1}{c|}{22.85}  & 28.50   & \multicolumn{1}{c|}{88.57}   & 54.13  &  11.56     \\
\multicolumn{1}{l|}{Qwen2.5-72B}    & \textbf{95.86}  & 62.53  & 35.55  & \multicolumn{1}{c|}{33.85}  & 35.75   & \multicolumn{1}{c|}{94.69}   & 41.05  &  13.95     \\
\multicolumn{1}{l|}{Qwen3-32B}  & 93.32& 86.14& \textbf{98.50} & \multicolumn{1}{c|}{54.75}  & 56.70   & \multicolumn{1}{c|}{96.52}   & 93.75  & 50.72     \\
\multicolumn{1}{l|}{Qwen3-30B-A3B}           & 93.14  & 85.20  & 98.15  & \multicolumn{1}{c|}{43.89}  & 58.25   & \multicolumn{1}{c|}{75.36}   & 92.85  &    44.11     \\
\multicolumn{1}{l|}{Llama3.1-8B}             & 88.58  & 9.05   & 62.15  & \multicolumn{1}{c|}{3.30}   & 8.80    & \multicolumn{1}{c|}{37.46}   & 45.15  &  2.74       \\
\multicolumn{1}{l|}{Llama3.3-70B}            & 94.74  & 49.26  & 94.72  & \multicolumn{1}{c|}{40.30}  & 41.75   & \multicolumn{1}{c|}{96.53}   & 88.95  &  37.25       \\
\multicolumn{1}{l|}{Llama4-Scout-17B-16E}    & 94.85  & 62.65  & 97.11  & \multicolumn{1}{c|}{29.04}  & 49.50    & \multicolumn{1}{c|}{58.67}   & 91.24  &   29.54     \\
\multicolumn{1}{l|}{Mistral-Small-3.1-24B}   & 94.65  & 28.96  & 49.55  & \multicolumn{1}{c|}{24.55}  & 36.65   & \multicolumn{1}{c|}{66.98}   & 77.55  &   16.87     \\
\multicolumn{1}{l|}{Mistral-Large-123B}      & 94.85  & 48.95  & 43.74  & \multicolumn{1}{c|}{43.12}  & 50.61 & \multicolumn{1}{c|}{81.62}   & 62.25  &   21.85     \\ 
\multicolumn{1}{l|}{DeepSeek-V3}             & 95.12  & 65.25  & 98.35  & \multicolumn{1}{c|}{68.65}  & \textbf{71.10}   & \multicolumn{1}{c|}{\textbf{96.55}}   & 93.43  &   45.74      \\
\multicolumn{1}{l|}{DeepSeek-R1}    & 94.92  & \textbf{93.61}  & 98.15  & \multicolumn{1}{c|}{\textbf{62.50}}  & 69.80   & \multicolumn{1}{c|}{89.54}   & \textbf{97.92}  &   \textbf{57.14}      \\\hline
\multicolumn{9}{c}{Closed-source models}                                                                                                                          \\ \hline
\multicolumn{1}{l|}{GPT-4o}                  & 95.22  & 64.45  & 45.64  & \multicolumn{1}{c|}{42.60}  & 45.30   & \multicolumn{1}{c|}{\textbf{94.04}}   & 57.65  &  24.87    \\
\multicolumn{1}{l|}{GPT-4.1}        & \textbf{95.30}  & 73.29  & 45.71  & \multicolumn{1}{c|}{59.90}  & 63.80   & \multicolumn{1}{c|}{93.89}   & 59.31  &   31.74      \\
\multicolumn{1}{l|}{o4-mini}                 & 95.25  & 90.72  & 98.25  & \multicolumn{1}{c|}{55.05}  & 64.61   & \multicolumn{1}{c|}{85.22}   & 93.75  &  56.23        \\
\multicolumn{1}{l|}{o3}             & 95.02  & 89.12  & 98.45  & \multicolumn{1}{c|}{\textbf{71.51}}  & \textbf{78.62}   & \multicolumn{1}{c|}{90.97}   & 91.44  &   \textbf{64.87}     \\
\multicolumn{1}{l|}{GPT-5}             & 95.18  & \textbf{93.40}  & \textbf{98.52}  & \multicolumn{1}{c|}{66.85}  & 71.65   & \multicolumn{1}{c|}{93.30}   & \textbf{94.45}  &   61.24\\
\multicolumn{1}{l|}{Claude3.5-Sonnet}        & 94.75  & 80.85  & 97.61  & \multicolumn{1}{c|}{33.05}  & 65.85   & \multicolumn{1}{c|}{50.19}   & 92.61  &  33.04       \\
\multicolumn{1}{l|}{Claude3.7-Sonnet}        & 94.85  & 74.18  & 98.11  & \multicolumn{1}{c|}{57.43}  & 66.35   & \multicolumn{1}{c|}{86.51}   & 93.53  &   57.91      \\
\multicolumn{1}{l|}{Gemini2.5-Flash-Preview} & 94.69  & 89.72  & 98.12  & \multicolumn{1}{c|}{35.85}  & 50.35   & \multicolumn{1}{c|}{71.20}   & 93.33  &   38.69      \\
\multicolumn{1}{l|}{Gemini2.5-Pro-Preview}   & 94.86  & 91.45  & 97.95  & \multicolumn{1}{c|}{54.80}  & 76.90   & \multicolumn{1}{c|}{71.26}   & 94.15  &   56.25      \\ 
\bottomrule[1.5pt]
\end{tabular}}
\caption{The performance of open and closed-source models on the SMART benchmark.}

\label{tab.smartresult}

\end{table*}

\subsubsection{Data Curation}
As shown in Tab.\ref{tab.sum_smart}, the questions used in the Understanding and Reasoning tasks, as well as the ground truths for the Reasoning, Arithmetic, and Refinement tasks, are directly derived from the original seed questions and therefore do not require additional verification. This design significantly reduces the cost of human annotation while maintaining high data quality. Below, we describe the curation process for the remaining dimensions.

\noindent \textbf{Structured Key Information.} The ground-truth for the Understanding task is the structured key information, which is generated by GPT-4.1.

\noindent \textbf{Notation-based Questions.} For the Arithmetic task, the input questions are notation-based questions. Directly converting a seed question into a notation problem is non-trivial, as it requires simplifying natural language into structured mathematical operations while preserving logical relationships among variables. To address this challenge, we adopt a two-stage process: seed questions are first formalized into SMT-LIB representations using GPT-4.1 to capture their underlying logic, and these formal expressions are subsequently translated into notation-based arithmetic questions also with GPT-4.1.

\noindent \textbf{R\&R.} For the Reflection \& Refinement (R\&R) task, the input question consists of the seed question paired with a chain-of-thought (CoT) solution containing deliberately injected errors. The outputs are the error categories and the corrected CoT. We define six error categories: arithmetic inaccuracies, omitted steps, hallucinated content, logical disorder, redundancy, and operator misuse. Error CoTs and their error types are generated according to predefined rules, so no additional verification is required. Detailed descriptions of these error types are provided in the Appendix Fig.\ref{fig.cotmistaks2}.

\subsubsection{Quality Control}
\label{sec:dataset_verification}
In Fig.~\ref{fig:pipline}, only the ground-truth for Understanding task and the questions for Arithmetic task are generated by LLMs, and thus require additional verification. To ensure the quality of the SMART benchmark, we implement a combined neuro-symbolic method and human verification procedure. This mechanism identifies and filters out low-quality samples and iteratively regenerates new data until the required quality standards are met.

\noindent \textbf{Neuro-symbolic Verification.}
We employ neuro-symbolic verification to ensure the correctness of the SMT-LIB expressions used in the Arithmetic dimension. Directly comparing the generated SMT-LIB with ground-truth expressions is challenging. Instead, we leverage the Z3 Solver~\cite{de2008z3} to automatically compute the result of a symbolic formula and compare it against the ground-truth answer from the original seed question, as a correct SMT-LIB expression should yield the same answer as its seed question.  This generation–validation process is repeated until the SMT-LIB expression produces the correct answer. 
 
\noindent \textbf{Human Verification.} The notation-base questions as question for the Arithmetic task and the structured key information as ground-truth for the Understanding task are both performed using GPT-4.1 and cannot be validated by the neuro-symbolic method. Thus, we follow the human verification protocol proposed by \cite{chen2024dr} to ensure the reliability of these generated data. Specifically, a randomly selected 10\% subset is manually reviewed by human annotators. If the sampled data fail to meet quality standards, the low-quality portions are regenerated, then re-sampled and re-verified until they pass inspection.

As a result, SMART benchmark comprises 10,000 test instances, including 2,000 original seed questions and 8,000 carefully curated, task-specific variants. Through a combination of neuro-symbolic methods and human verification, we ensure that each instance meets quality standards. The main differences to existing benchmarks are discussed in the Append. \ref{append.dif}
\section{Experiments}
\subsection{Models}
We evaluate 22 recent open- and closed-source LLMs using our SMART framework, covering both general-purpose and reasoning-specialized models. To ensure deterministic outputs, the temperature is set to 0. For each dimension-specific task, we employ a three-shot prompting strategy.

\begin{figure*}[t]
\centering
  \includegraphics[width=\linewidth]{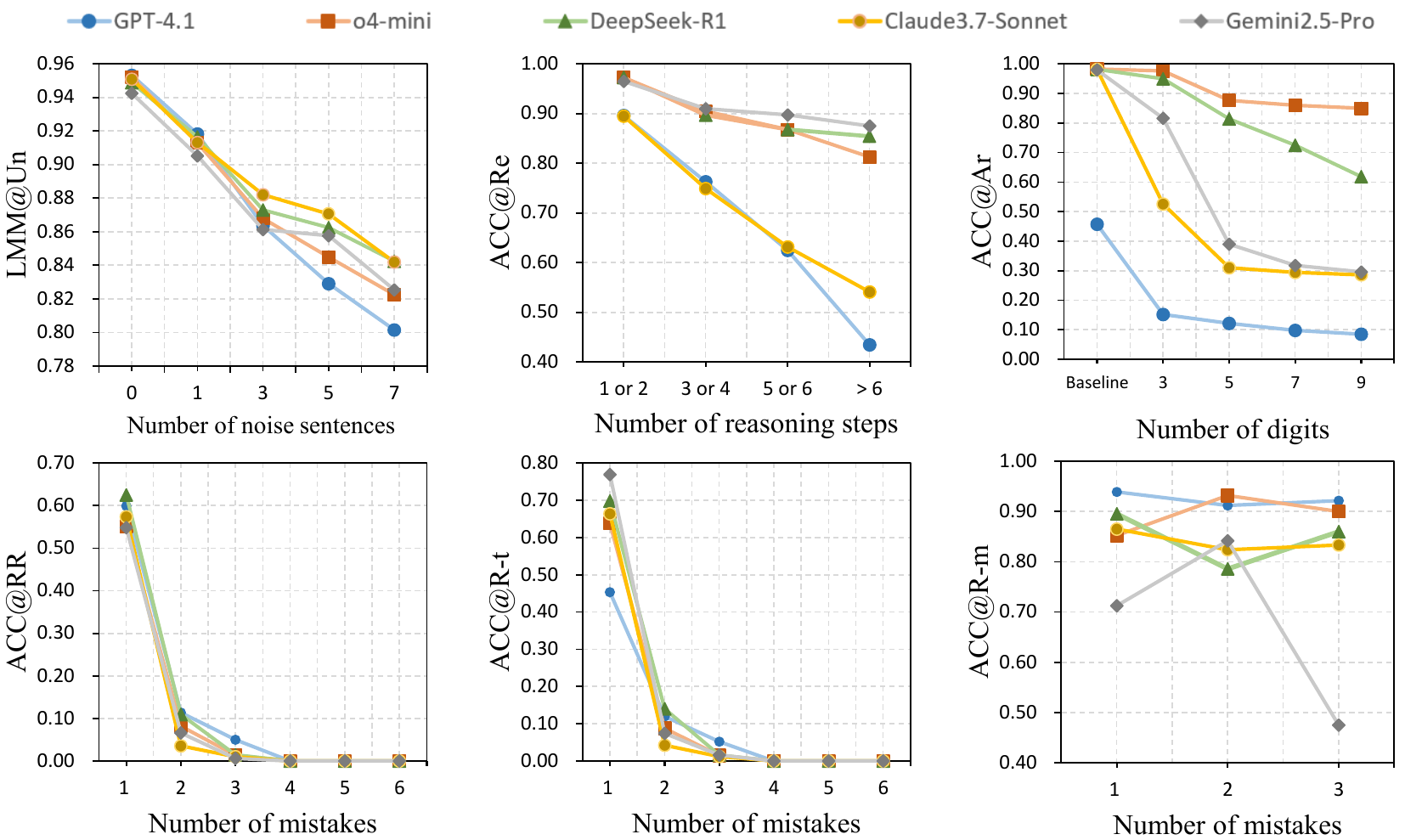}
  \caption{The performance across the varying difficulty settings for each SMART dimension.}

  \label{fig:diff}

\end{figure*}
\subsection{Evaluation Metrics}
\noindent \textbf{Understanding.} We adopt the LLM-as-a-Judge evaluation approach \cite{zheng2023judging}, introducing the metric LLM@Un to evaluate the quality of generated structured information. To mitigate potential preference bias~\cite{li2502preference} inherent in LLM-based judging, we independently employ GPT-4.1 and DeepSeek-V3 as two judging models. Each judge evaluates the semantic similarity between the model-generated context and the reference ground-truth, assigning a similarity score ranging from 1 to 100. The final LLM@Un score is the average of the scores from GPT-4.1 and DeepSeek-V3.

\noindent \textbf{Reasoning.}  We do not evaluate the correctness of the generated SMT-LIB expressions, since multiple logically equivalent solutions can exist for a problem. Instead, we validate these symbolic expressions by executing them with the Z3 Solver and evaluating the solver’s output with the ground-truth answer of the seed question via rule-matching. The metric for the Arithmetic task is accuracy and is computed as $ACC@Re = \frac{N_{correct}}{N_{total}}$, which means the percentage of accurately answered questions.

\noindent \textbf{Arithmetic.} We evaluate answers of notation-based questions by comparing them with the ground-truth of the seed questions using rule matching, and use accuracy-based metrics ACC@Ar.

\noindent \textbf{R\&R.} For the Reflection task, models are required to identify error categories within a given CoT. The outputs are compared with the ground-truth categories, and the accuracy-based metric is ACC@R-t. For the Refinement task, the final answer is extracted from the refined CoT using rules. The refined solution is considered correct if the extracted answer matches the ground-truth of the seed question, and use accuracy-based metric ACC@R-m.

\noindent \textbf{All-Pass Score.} The All-Pass Score is an integrated metric (ACC@All) that combines performance across all evaluation dimensions. Specifically, a model achieves an All-Pass success if it simultaneously meets the following criteria: (1) obtaining a score of at least 90\% on the Understanding task; (2) correctly solving the Reasoning task; (3) correctly solving the Arithmetic task; and (4) successfully completing the entire R\&R task. We require at least 90\%  on Understanding to demand near-exact semantic extraction while allowing minimal lexical variation, which stabilizes LLM-as-a-Judge scoring and keeps All-Pass difficulty comparable across dimensions.

\begin{table*}[]
\resizebox{\textwidth}{!}{%
\setlength{\tabcolsep}{5pt}
\centering
\renewcommand\arraystretch{1.1}
\begin{tabular}{l|l|cc|cccccccc}
\toprule[1.5pt]
\multirow{2}{*}{\rotatebox{90}{Model}} &
  \multirow{2}{*}{Perturbation} &
  \multicolumn{2}{c|}{Final Answer} &
  \multicolumn{2}{c}{Understanding} &
  \multicolumn{2}{c}{Reasoning} &
  \multicolumn{2}{c}{Arithmetic} &
  \multicolumn{2}{c}{R\&R}\\ \cline{3-12} 
                              &                       & ACC@An & PD   & LLM@Un   & PD    & ACC@Reason & PD    & ACC@Ar  & PD &ACC@RR &PD   \\ \hline
\multirow{4}{*}{\rotatebox{90}{GPT-4.1}} & Seed question         & 59.31 & /       & 95.41  & /       & 73.29    & /       & 45.71 & /   &59.90 & /    \\
                              & + Noise insertion     &  41.23& 18.08$\downarrow$ & 88.52 &6.89$\downarrow$ & 46.54  &\textbf{26.75}$\downarrow$  &39.40  &6.31$\downarrow$ &45.67&14.23$\downarrow$ \\
                              & + Adding operation    &28.45  &30.86$\downarrow$  & 91.32 &4.09$\downarrow$ & 57.53 & 15.76$\downarrow$& 20.64 & \textbf{25.07}$\downarrow$&48.12&11.78$\downarrow$ \\
                              & + Numerical variation & 24.81 &34.50$\downarrow$  & 92.83 & 2.58$\downarrow$ &52.81  &20.48$\downarrow$  & 18.34 & \textbf{27.37}$\downarrow$&51.86&8.04$\downarrow$\\ \hline
\multirow{4}{*}{\rotatebox{90}{Claude-3.7}}        & Seed question         & 93.53 & /       &95.09 & /       & 74.18    & /       & 98.11 & /  &57.43 &/     \\
                              & + Noise insertion     &87.60  &5.93$\downarrow$ &89.87  & 5.22$\downarrow$&65.81  &\textbf{8.37}$\downarrow$ &91.82 &  6.29$\downarrow$&47.85&9.58$\downarrow$\\
                              & + Adding operation    &76.21  & 17.32$\downarrow$&92.34  &2.75$\downarrow$& 54.38  & 19.80$\downarrow$&67.43 & \textbf{24.39}$\downarrow$ &50.78&6.65$\downarrow$  \\
                              & + Numerical variation &  62.74 & 30.79$\downarrow$ & 93.17  & 1.92$\downarrow$ &   58.92  &15.26$\downarrow$ & 58.29 &  \textbf{39.82}$\downarrow$  &52.85&4.58$\downarrow$   \\ \bottomrule[1.5pt]
\end{tabular}}
\caption{The performance degradation of evaluation dimensions when three types of perturbations are added to the seed questions. PD refers to the performance drop. The most affected dimension in each case is highlighted in bold.}

\label{tab.pert}
\end{table*}

\begin{table}[]
\resizebox{\columnwidth}{!}{%
\renewcommand\arraystretch{1.1}
\begin{tabular}{l|ccc}
\toprule[1.5pt]
Method & \begin{tabular}[c]{@{}c@{}}Refined\\  Dimension\end{tabular}  & ACC@Fi & \begin{tabular}[c]{@{}c@{}}ACC@Fi\\   with self-refine\end{tabular} \\ \hline
Llama3.1-8B   & Reasoning & 45.15 & 43.85 ($\downarrow1.30$) \\
Mistral-Small-24B & Reasoning & 77.55 &78.15 ($\uparrow0.61$)\\
GLM-24B & Arithmetic & 54.13 & 56.28 ($\uparrow2.15$) \\
Qwen2.5-72B & Arithmetic & 41.05 & 52.82 ($\uparrow11.77$) \\ \bottomrule[1.5pt]
\end{tabular}}
\caption{Self-Refine prompting on a specific dimension.}
\label{tab.selfprompt}

\end{table}

\subsection{Performance on the SMART benchmark}
The performance of the 22 evaluated LLMs on the SMART benchmark is detailed in Tab.~\ref{tab.smartresult},  which reports the scores for all evaluation dimensions, All-Pass Score, and final answer accuracy.

\subsubsection{Dimension specific-task results} Our results indicate that the LLMs generally demonstrate a strong capacity for problem understanding, with most LLM@Un scores exceeding 90\%. This suggests a general proficiency in grasping relevant information and interpreting problem statements. However, significant performance disparities emerge in the Reasoning dimension, where ACC@Re scores range widely from 9.05\% to 93.61\%. A similar divergence is observed in the Reflection task, with the highest score (78.62\%) being nearly nine times greater than the lowest (8.80\%). These findings suggest that symbolic reasoning and error reflection capabilities represent critical bottlenecks, particularly for smaller models. In contrast, the Arithmetic and Refinement tasks appear relatively less challenging, with leading LLMs achieving near-perfect performance. For example, o3 attains an ACC@Ar of 98.45\%, while DeepSeek-V3 reaches an ACC@R-m of 96.55\%, demonstrating their strength in computational and corrective capabilities.

\subsubsection{Granular insights} The SMART benchmark framework uncovers nuanced performance differences among LLMs that are obscured by final answer metrics alone. For example, while o4-mini and Claude3.7-Sonnet exhibit similar final answer accuracies (93.75\% and 93.53\%, respectively), o4-mini demonstrates markedly higher proficiency in the Reasoning dimension (90.72\%) compared to Claude3.7-Sonnet (74.18\%). A similar trend is observed when comparing o4-mini and DeepSeek-V3, further illustrating SMART benchmark’s capability to reveal fine-grained gaps that traditional outcome-based metrics miss. Additionally, although o4-mini and GPT-4.1 perform similarly on the Understanding and Reflection \& Refinement dimensions, o4-mini’s final answer accuracy (93.75\%) is markedly higher than GPT-4.1’s (59.31\%). 
Our framework attributes this disparity primarily to GPT-4.1's lower capability in two key dimensions—Reasoning, where it scored 73.29\% compared to o4-mini’s 90.72\%, and Arithmetic, where it achieved only 45.71\% in contrast to o4-mini’s 98.25\%. Thus, the SMART framework facilitates a deeper analysis and interpretation of the underlying causes for performance differences.

\subsubsection{All-Pass Score remains a challenge} The All-Pass Score serves as a rigorous discriminator of model capability. The top-performing model, o3, achieves only 64.87\% on this metric, significantly lagging behind its final answer accuracy of 91.44\%. This disparity reveals that models often fail in specific cognitive dimensions even when the final answer is correct. The All-Pass Score confirms that SMART remains a challenging benchmark with substantial room for improvement.
\subsection{How Does Task Difficulty Impact Different Dimensions of SMART?}
\label{sec.dysmart1}
To investigate how task difficulty affects model performance across different SMART dimensions, we construct new sets of dimension-specific questions with varied difficulty levels and evaluate five leading closed-source LLMs on this dynamic test set. Task difficulty is manipulated in the following ways: for the Understanding dimension, by varying the number of added irrelevant sentences; for the Reasoning dimension, by grouping questions according to the number of required reasoning steps; for the Arithmetic dimension, by changing the number of digits (referring to digit length in scientific notation, not the number of operands); and for the R\&R dimension, by altering the mistakes introduced into the CoT. 

It is important to note that modifying the number of digits in arithmetic questions changes the ground-truth answer. To ensure correctness, we simultaneously update both the numerical values in the arithmetic questions and their corresponding SMT-LIB representations, subsequently employing the Z3 Solver to generate new ground-truth answers. For the other dimensions, the ground-truth answers remain unchanged. 

As shown in Fig.~\ref{fig:diff}, increasing task complexity generally leads to notable performance degradation across all dimensions. Notably, GPT-4.1 and Claude3.7-Sonnet show pronounced sensitivity in the Reasoning dimension, with ACC@Re scores dropping sharply from approximately 90\% to below 60\% as the number of reasoning steps increases. In contrast, the remaining models maintain ACC@Re scores above 80\% even with more than six reasoning steps. In the Arithmetic dimension, o4-mini demonstrates robust performance even with nine-digit numbers, whereas GPT-4.1's accuracy falls below 10\%. For Reflection tasks, introducing just two error types into the CoT results in a steep decline in detection accuracy for all models, with none able to reliably detect all errors when four or more distinct mistake types are present.

\subsection{How Do Fine-grained Dimensions Influence the Performance of Final Answer Accuracy?}
\label{sec.dysmart2}
Prior work has shown that LLMs experience significant drops in final answer accuracy when evaluated on perturbed versions of questions~\cite{li2024gsm, zhu2023dyval, li2024perteval}. However, the underlying reasons for this degradation remain insufficiently explored. To address this gap, we adapt three perturbation strategies from~\cite{li2024gsm} and apply them to both the seed questions and their corresponding dimension-specific variants, aiming to identify which dimensions are most susceptible to performance loss under these perturbations.

As shown in Tab.~\ref{tab.pert}, all evaluated dimensions exhibit substantial performance drops (PD) under perturbations. When noise is introduced, the reasoning dimension is most affected for both GPT-4.1 (26.75\%) and Claude 3.7-Sonnet (8.37\%). Conversely, additional operations or numerical modifications lead to the greatest drops in the Arithmetic dimension. These results suggest that irrelevant information primarily undermines reasoning capabilities, while changes to operations or numeric values predominantly impact arithmetic proficiency. Ultimately, vulnerabilities across all dimensions collectively reduce final answer accuracy.

\subsection{Improving LLMs via Self-Refine Prompting on Weak Dimensions}
To enhance the mathematical capabilities of LLMs, we apply self-refinement prompting specifically to the weakest step identified in SMART. The specific prompts are provided in the Appendix. As shown in Tab.~\ref{tab.selfprompt}, targeting the reasoning or arithmetic dimension leads to notable performance gains for Gemma3-27B, Mistral-Small, and Qwen2.5-72B. In contrast, Llama3.1-8B experiences a slight performance drop, likely due to its limited capacity for self-reflection. These results demonstrate that the SMART framework is an effective diagnostic tool for pinpointing a model’s weakest dimension and that targeted intervention on this dimension can improve mathematical performance.

\section{Conclusion}
We present SMART, a benchmark designed to evaluate the mathematical problem-solving capabilities of LLMs. Inspired by Pólya’s theory of problem solving, SMART decomposes the reasoning process into four cognitive dimensions—Understanding, Reasoning, Arithmetic, and Reflection \& Refinement—and introduces a novel All-Pass Score metric for comprehensive evaluation. We also propose a data curation and quality control framework that iteratively verifies generated test data to ensure reliability. Experiments on 22 open- and closed-source LLMs reveal that Reasoning and Reflection remain key bottlenecks, while targeted improvements on weak dimensions can enhance overall mathematical capability. We hope SMART provides a foundation for more systematic and interpretable evaluation of LLMs’ reasoning processes in future research.

\section{Limitation}

While our proposed framework provides a comprehensive evaluation platform, it is important to acknowledge its scope limitations. In particular, although Z3 and SMT-LIB effectively handle linear, integer, and some nonlinear constraints, their problem-solving capabilities are restricted. They are unsuitable for highly complex nonlinear problems and certain NP-complete combinatorial tasks. Whether SMART targets algebraic questions or more advanced domains is determined by the choice of formal language and prover, rather than by the SMART framework itself. To overcome these limitations, future work will investigate using Lean~\cite{moura2021lean} to formalize and prove complex mathematical theorems involving higher-order logic and intricate proof structures beyond SMT solvers' scope.
\section{Acknowledgements}
This work was supported by the National Natural Science
Foundation of China (62437001 and 62402051) and the Fundamental Research Funds for the Central Universities  (2243100020 and 225310002). 
\bibliography{custom}

@article{cobbe2021training,
  title={Training verifiers to solve math word problems},
  author={Cobbe, Karl and Kosaraju, Vineet and Bavarian, Mohammad and Chen, Mark and Jun, Heewoo and Kaiser, Lukasz and Plappert, Matthias and Tworek, Jerry and Hilton, Jacob and Nakano, Reiichiro and others},
  journal={arXiv preprint arXiv:2110.14168},
  year={2021}
}

@inproceedings{ling2017program,
  title={Program Induction by Rationale Generation: Learning to Solve and Explain Algebraic Word Problems},
  author={Ling, Wang and Yogatama, Dani and Dyer, Chris and Blunsom, Phil},
  booktitle={Proceedings of the 55th Annual Meeting of the Association for Computational Linguistics (Volume 1: Long Papers)},
  pages={158--167},
  year={2017}
}

@inproceedings{patel2021nlp,
  title={Are NLP Models really able to Solve Simple Math Word Problems?},
  author={Patel, Arkil and Bhattamishra, Satwik and Goyal, Navin},
  booktitle={Proceedings of the 2021 Conference of the North American Chapter of the Association for Computational Linguistics: Human Language Technologies},
  pages={2080--2094},
  year={2021}
}

@inproceedings{miao2020diverse,
  title={A Diverse Corpus for Evaluating and Developing English Math Word Problem Solvers},
  author={Miao, Shen-Yun and Liang, Chao-Chun and Su, Keh-Yih},
  booktitle={Proceedings of the 58th Annual Meeting of the Association for Computational Linguistics},
  pages={975--984},
  year={2020}
}

@inproceedings{koncel2016mawps,
  title={MAWPS: A math word problem repository},
  author={Koncel-Kedziorski, Rik and Roy, Subhro and Amini, Aida and Kushman, Nate and Hajishirzi, Hannaneh},
  booktitle={Proceedings of the 2016 conference of the north american chapter of the association for computational linguistics: human language technologies},
  pages={1152--1157},
  year={2016}
}

@inproceedings{li2024gsm,
  title={GSM-Plus: A Comprehensive Benchmark for Evaluating the Robustness of LLMs as Mathematical Problem Solvers},
  author={Li, Qintong and Cui, Leyang and Zhao, Xueliang and Kong, Lingpeng and Bi, Wei},
  booktitle={Proceedings of the 62nd Annual Meeting of the Association for Computational Linguistics (Volume 1: Long Papers)},
  pages={2961--2984},
  year={2024}
}

@inproceedings{mirzadehgsm,
  title={GSM-Symbolic: Understanding the Limitations of Mathematical Reasoning in Large Language Models},
  author={Mirzadeh, Seyed Iman and Alizadeh, Keivan and Shahrokhi, Hooman and Tuzel, Oncel and Bengio, Samy and Farajtabar, Mehrdad},
  booktitle={The Thirteenth International Conference on Learning Representations},
  year={2025}

}

@inproceedings{zhu2023dyval,
  title={Dyval: Dynamic evaluation of large language models for reasoning tasks},
  author={Zhu, Kaijie and Chen, Jiaao and Wang, Jindong and Gong, Neil Zhenqiang and Yang, Diyi and Xie, Xing},
  booktitle={The Twelfth International Conference on Learning Representations},
  year={2023}
}

@article{achiam2023gpt,
  title={Gpt-4 technical report},
  author={Achiam, Josh and Adler, Steven and Agarwal, Sandhini and Ahmad, Lama and Akkaya, Ilge and Aleman, Florencia Leoni and Almeida, Diogo and Altenschmidt, Janko and Altman, Sam and Anadkat, Shyamal and others},
  journal={arXiv preprint arXiv:2303.08774},
  year={2023}
}

@article{OpenAI2024,
    author    = {OpenAI},
    title     = {Learning to reason with LLMs},
    year      = {2024},
    url       = {https://openai.com/index/learning-to-reason-with-llms/},

}

@article{OpenAIo3-o4mini,
    author    = {OpenAI},
    title     = {Introducing OpenAI o3 and o4-mini},
    year      = {2025},
    url       = {https://openai.com/index/introducing-o3-and-o4-mini/},

}

@article{OpenAIgpt4.1,
    author    = {OpenAI},
    title     = {Introducing GPT-4.1 in the API},
    year      = {2025},
    url       = {https://openai.com/index/gpt-4-1/},

}

@article{gemin25,
    author    = {Google},
    title     = {Gemini 2.5: Our most intelligent AI model},
    year      = {2024},
    url       = {https://blog.google/technology/google-deepmind/gemini-model-thinking-updates-march-2025/},
}

@article{claude35,
    author    = {anthropic},
    title     = {Claude 3.5 Sonnet},
    year      = {2024},
    url       = {https://www.anthropic.com/news/claude-3-5-sonnet},
}

@article{claude37,
    author    = {anthropic},
    title     = {Claude 3.7 Sonnet and Claude Code},
    year      = {2025},
    url       = {https://www.anthropic.com/news/claude-3-7-sonnet},
}

@inproceedings{zhu2024dynamic,
  title={Dynamic Evaluation of Large Language Models by Meta Probing Agents},
  author={Zhu, Kaijie and Wang, Jindong and Zhao, Qinlin and Xu, Ruochen and Xie, Xing},
  booktitle={International Conference on Machine Learning},
  pages={62599--62617},
  year={2024},
  organization={PMLR}
}

@inproceedings{barrett2010smt,
  title={The smt-lib standard: Version 2.0},
  author={Barrett, Clark and Stump, Aaron and Tinelli, Cesare and others},
  booktitle={Proceedings of the 8th international workshop on satisfiability modulo theories (Edinburgh, UK)},
  volume={13},
  pages={14},
  year={2010}
}

@inproceedings{de2008z3,
  title={Z3: An efficient SMT solver},
  author={De Moura, Leonardo and Bj{\o}rner, Nikolaj},
  booktitle={International conference on Tools and Algorithms for the Construction and Analysis of Systems},
  pages={337--340},
  year={2008},
  organization={Springer}
}

@book{polya2014solve,
  title={How to solve it: A new aspect of mathematical method},
  author={Polya, George},
  volume={34},
  year={2014},
  publisher={Princeton university press}
}

@book{polya1957solve,
  title={How to solve it: A new aspect of mathematical method},
  author={P{\'o}lya, George and Conway, John Horton},
  year={1957},
  publisher={Princeton University Press Princeton}
}

@misc{llama4,
  author = {Meta},
  title = {llama4-MODEL-CARD.md},
  year = {2025},
  publisher = {GitHub},
  journal = {GitHub repository},
  howpublished = {\url{https://github.com/meta-llama/llama-models/blob/main/models/llama4/MODEL_CARD.md}},
}

@article{wei2022chain,
  title={Chain-of-thought prompting elicits reasoning in large language models},
  author={Wei, Jason and Wang, Xuezhi and Schuurmans, Dale and Bosma, Maarten and Xia, Fei and Chi, Ed and Le, Quoc V and Zhou, Denny and others},
  journal={Advances in neural information processing systems},
  volume={35},
  pages={24824--24837},
  year={2022}
}

@misc{mistral-small,
  author = {MistralAITeam},
  title = {Mistral-Small-Instruct-2409},
  year = {2024},
  howpublished = {\url{https://huggingface.co/mistralai/Mistral-Small-Instruct-2409}},
}

@misc{qwen2.5,
    title = {Qwen2.5: A Party of Foundation Models},
    url = {https://qwenlm.github.io/blog/qwen2.5/},
    author = {Qwen Team},
    month = {September},
    year = {2024}
}

@article{abdin2024phi,
  title={Phi-4 technical report},
  author={Abdin, Marah and Aneja, Jyoti and Behl, Harkirat and Bubeck, S{\'e}bastien and Eldan, Ronen and Gunasekar, Suriya and Harrison, Michael and Hewett, Russell J and Javaheripi, Mojan and Kauffmann, Piero and others},
  journal={arXiv preprint arXiv:2412.08905},
  year={2024}
}

@article{llama3modelcard,
title={Llama 3 Model Card},
author={AI@Meta},
year={2024},
url = {https://github.com/meta-llama/llama3/blob/main/MODEL_CARD.md}
}

@article{liu2024deepseek,
  title={Deepseek-v3 technical report},
  author={Liu, Aixin and Feng, Bei and Xue, Bing and Wang, Bingxuan and Wu, Bochao and Lu, Chengda and Zhao, Chenggang and Deng, Chengqi and Zhang, Chenyu and Ruan, Chong and others},
  journal={arXiv preprint arXiv:2412.19437},
  year={2024}
}

@article{wang2024large,
  title={Large language models for education: A survey and outlook},
  author={Wang, Shen and Xu, Tianlong and Li, Hang and Zhang, Chaoli and Liang, Joleen and Tang, Jiliang and Yu, Philip S and Wen, Qingsong},
  journal={arXiv preprint arXiv:2403.18105},
  year={2024}
}

@inproceedings{hendrycks2measuring,
  title={Measuring Mathematical Problem Solving With the MATH Dataset},
  author={Hendrycks, Dan and Burns, Collin and Kadavath, Saurav and Arora, Akul and Basart, Steven and Tang, Eric and Song, Dawn and Steinhardt, Jacob},
  booktitle={Thirty-fifth Conference on Neural Information Processing Systems Datasets and Benchmarks Track (Round 2)},
  year={2021}

}

@article{glazer2024frontiermath,
  title={Frontiermath: A benchmark for evaluating advanced mathematical reasoning in ai},
  author={Glazer, Elliot and Erdil, Ege and Besiroglu, Tamay and Chicharro, Diego and Chen, Evan and Gunning, Alex and Olsson, Caroline Falkman and Denain, Jean-Stanislas and Ho, Anson and Santos, Emily de Oliveira and others},
  journal={arXiv preprint arXiv:2411.04872},
  year={2024}
}

@inproceedings{moura2021lean,
  title={The Lean 4 theorem prover and programming language},
  author={Moura, Leonardo de and Ullrich, Sebastian},
  booktitle={Automated Deduction--CADE 28: 28th International Conference on Automated Deduction, Virtual Event, July 12--15, 2021, Proceedings 28},
  pages={625--635},
  year={2021},
  organization={Springer}
}

@article{aime2024,
title={aime2024},
author={Huggingface},
year={2024},
url = {https://huggingface.co/datasets/Maxwell-Jia/AIME_2024}
}

@article{team2025gemma,
  title={Gemma 3 technical report},
  author={Team, Gemma and Kamath, Aishwarya and Ferret, Johan and Pathak, Shreya and Vieillard, Nino and Merhej, Ramona and Perrin, Sarah and Matejovicova, Tatiana and Ram{\'e}, Alexandre and Rivi{\`e}re, Morgane and others},
  journal={arXiv preprint arXiv:2503.19786},
  year={2025}
}

@article{glm2024chatglm,
  title={Chatglm: A family of large language models from glm-130b to glm-4 all tools},
  author={GLM, Team and Zeng, Aohan and Xu, Bin and Wang, Bowen and Zhang, Chenhui and Yin, Da and Zhang, Dan and Rojas, Diego and Feng, Guanyu and Zhao, Hanlin and others},
  journal={arXiv preprint arXiv:2406.12793},
  year={2024}
}

@article{yang2025qwen3,
  title={Qwen3 Technical Report},
  author={Yang, An and Li, Anfeng and Yang, Baosong and Zhang, Beichen and Hui, Binyuan and Zheng, Bo and Yu, Bowen and Gao, Chang and Huang, Chengen and Lv, Chenxu and others},
  journal={arXiv preprint arXiv:2505.09388},
  year={2025}
}

@article{guo2025deepseek,
  title={Deepseek-r1: Incentivizing reasoning capability in llms via reinforcement learning},
  author={Guo, Daya and Yang, Dejian and Zhang, Haowei and Song, Junxiao and Zhang, Ruoyu and Xu, Runxin and Zhu, Qihao and Ma, Shirong and Wang, Peiyi and Bi, Xiao and others},
  journal={arXiv preprint arXiv:2501.12948},
  year={2025}
}

@article{zheng2023judging,
  title={Judging llm-as-a-judge with mt-bench and chatbot arena},
  author={Zheng, Lianmin and Chiang, Wei-Lin and Sheng, Ying and Zhuang, Siyuan and Wu, Zhanghao and Zhuang, Yonghao and Lin, Zi and Li, Zhuohan and Li, Dacheng and Xing, Eric and others},
  journal={Advances in Neural Information Processing Systems},
  volume={36},
  pages={46595--46623},
  year={2023}
}

@inproceedings{ma-etal-2025-problem,
    title = "Problem-Solving Logic Guided Curriculum In-Context Learning for {LLM}s Complex Reasoning",
    author = "Ma, Xuetao  and
      Jiang, Wenbin  and
      Huang, Hua",
    editor = "Che, Wanxiang  and
      Nabende, Joyce  and
      Shutova, Ekaterina  and
      Pilehvar, Mohammad Taher",
    booktitle = "Findings of the Association for Computational Linguistics: ACL 2025",
    month = jul,
    year = "2025",
    address = "Vienna, Austria",
    publisher = "Association for Computational Linguistics",
    url = "https://aclanthology.org/2025.findings-acl.440/",
    doi = "10.18653/v1/2025.findings-acl.440",
    pages = "8394--8412",
    ISBN = "979-8-89176-256-5"
}

@inproceedings{zengmr,
  title={MR-GSM8K: A Meta-Reasoning Benchmark for Large Language Model Evaluation},
  author={Zeng, Zhongshen and Chen, Pengguang and Liu, Shu and Jiang, Haiyun and Jia, Jiaya},
  booktitle={The Thirteenth International Conference on Learning Representations},
  year={2025}
}

@inproceedings{zhengminif2f,
  title={miniF2F: a cross-system benchmark for formal Olympiad-level mathematics},
  author={Zheng, Kunhao and Han, Jesse Michael and Polu, Stanislas},
  booktitle={International Conference on Learning Representations},
  year={2022}

}

@misc{liu2023fimo,
      title={FIMO: A Challenge Formal Dataset for Automated Theorem Proving}, 
      author={Chengwu Liu and Jianhao Shen and Huajian Xin and Zhengying Liu and Ye Yuan and Haiming Wang and Wei Ju and Chuanyang Zheng and Yichun Yin and Lin Li and Ming Zhang and Qun Liu},
      year={2023},
      eprint={2309.04295},
      archivePrefix={arXiv},
      primaryClass={cs.AI}
}

@inproceedings{song-etal-2025-prmbench,
    title = "{PRMB}ench: A Fine-grained and Challenging Benchmark for Process-Level Reward Models",
    author = "Song, Mingyang  and
      Su, Zhaochen  and
      Qu, Xiaoye  and
      Zhou, Jiawei  and
      Cheng, Yu",
    editor = "Che, Wanxiang  and
      Nabende, Joyce  and
      Shutova, Ekaterina  and
      Pilehvar, Mohammad Taher",
    booktitle = "Proceedings of the 63rd Annual Meeting of the Association for Computational Linguistics (Volume 1: Long Papers)",
    month = jul,
    year = "2025",
    address = "Vienna, Austria",
    publisher = "Association for Computational Linguistics",
    url = "https://aclanthology.org/2025.acl-long.1230/",
    doi = "10.18653/v1/2025.acl-long.1230",
    pages = "25299--25346",
    ISBN = "979-8-89176-251-0"
}

@article{li2502preference,
  title={Preference leakage: A contamination problem in llm-as-a-judge, 2025},
  author={Li, Dawei and Sun, Renliang and Huang, Yue and Zhong, Ming and Jiang, Bohan and Han, Jiawei and Zhang, Xiangliang and Wang, Wei and Liu, Huan},
  journal={URL https://arxiv. org/abs/2502.01534}
}

@article{zhang2024careful,
  title={A careful examination of large language model performance on grade school arithmetic},
  author={Zhang, Hugh and Da, Jeff and Lee, Dean and Robinson, Vaughn and Wu, Catherine and Song, William and Zhao, Tiffany and Raja, Pranav and Zhuang, Charlotte and Slack, Dylan and others},
  journal={Advances in Neural Information Processing Systems},
  volume={37},
  pages={46819--46836},
  year={2024}
}

@article{zeng2024mr,
  title={Mr-ben: A meta-reasoning benchmark for evaluating system-2 thinking in llms},
  author={Zeng, Zhongshen and Liu, Yinhong and Wan, Yingjia and Li, Jingyao and Chen, Pengguang and Dai, Jianbo and Yao, Yuxuan and Xu, Rongwu and Qi, Zehan and Zhao, Wanru and others},
  journal={Advances in Neural Information Processing Systems},
  volume={37},
  pages={119466--119546},
  year={2024}
}

@inproceedings{zheng-etal-2025-processbench,
    title = "{P}rocess{B}ench: Identifying Process Errors in Mathematical Reasoning",
    author = "Zheng, Chujie  and
      Zhang, Zhenru  and
      Zhang, Beichen  and
      Lin, Runji  and
      Lu, Keming  and
      Yu, Bowen  and
      Liu, Dayiheng  and
      Zhou, Jingren  and
      Lin, Junyang",
    editor = "Che, Wanxiang  and
      Nabende, Joyce  and
      Shutova, Ekaterina  and
      Pilehvar, Mohammad Taher",
    booktitle = "Proceedings of the 63rd Annual Meeting of the Association for Computational Linguistics (Volume 1: Long Papers)",
    month = jul,
    year = "2025",
    address = "Vienna, Austria",
    publisher = "Association for Computational Linguistics",
    url = "https://aclanthology.org/2025.acl-long.50/",
    doi = "10.18653/v1/2025.acl-long.50",
    pages = "1009--1024",
    ISBN = "979-8-89176-251-0"
}

@inproceedings{chen2024dr,
  title={Dr. Academy: A Benchmark for Evaluating Questioning Capability in Education for Large Language Models},
  author={Chen, Yuyan and Wu, Chenwei and Yan, Songzhou and Liu, Panjun and Xiao, Yanghua},
  booktitle={Proceedings of the 62nd Annual Meeting of the Association for Computational Linguistics (Volume 1: Long Papers)},
  pages={3138--3167},
  year={2024}
}

@article{li2024perteval,
  title={Perteval: Unveiling real knowledge capacity of llms with knowledge-invariant perturbations},
  author={Li, Jiatong and Hu, Renjun and Huang, Kunzhe and Zhuang, Yan and Liu, Qi and Zhu, Mengxiao and Shi, Xing and Lin, Wei},
  journal={Advances in Neural Information Processing Systems},
  volume={37},
  pages={10679--10706},
  year={2024}
}

@inproceedings{zhang2025ocr,
  title={Ocr hinders rag: Evaluating the cascading impact of ocr on retrieval-augmented generation},
  author={Zhang, Junyuan and Zhang, Qintong and Wang, Bin and Ouyang, Linke and Wen, Zichen and Li, Ying and Chow, Ka-Ho and He, Conghui and Zhang, Wentao},
  booktitle={Proceedings of the IEEE/CVF International Conference on Computer Vision},
  pages={17443--17453},
  year={2025}
}

@inproceedings{yang-etal-2025-well,
    title = "How Well Can Reasoning Models Identify and Recover from Unhelpful Thoughts?",
    author = "Yang, Sohee  and
      Lee, Sang-Woo  and
      Kassner, Nora  and
      Gottesman, Daniela  and
      Riedel, Sebastian  and
      Geva, Mor",
    editor = "Christodoulopoulos, Christos  and
      Chakraborty, Tanmoy  and
      Rose, Carolyn  and
      Peng, Violet",
    booktitle = "Findings of the Association for Computational Linguistics: EMNLP 2025",
    month = nov,
    year = "2025",
    address = "Suzhou, China",
    publisher = "Association for Computational Linguistics",
    url = "https://aclanthology.org/2025.findings-emnlp.370/",
    doi = "10.18653/v1/2025.findings-emnlp.370",
    pages = "7030--7047",
    ISBN = "979-8-89176-335-7",
}

@article{883d61ddf10640bfa3dfed182207b890,
title = "Numeracy, literacy and Newman's Error Analysis",
author = "White, {Allan Leslie}",
year = "2010",
language = "English",
volume = "33",
pages = "129--148",
journal = "Journal of Science and Mathematics Education in Southeast Asia",
issn = "0126-7663",
publisher = "Southeast Asian Ministers of Education Organisation: Regional Centre for Education in Science and Mathematics",
number = "2",
}

\appendix

\section{ Appendix}
\label{sec:appendix}

\subsection{P\'olya's Problem-Solving Theory}
P\'oya’s four-step problem-solving framework, proposed in the mid-20th century, has become a canonical model and is widely used to analyze students’ strategies and error patterns in mathematics education. Newman’s Error Analysis (NEA) \cite{883d61ddf10640bfa3dfed182207b890} decomposes students’ performance on mathematical word problems into sequential skills that closely mirror P\'olya’s stages, and is widely used to diagnose where in the problem-solving process students fail.

Following this line of work, the four evaluation dimensions in SMART are designed as an LLM-oriented realization of P\'olya’s theory and are directly aligned with human mathematical problem-solving processes.

Specifically, the Understanding task evaluates a model’s ability to extract and organize key information from the question. The Reasoning task evaluates the ability to devise a solution plan by producing a symbolic formalization. The Arithmetic task assesses the ability to carry out that plan by solving notation-based arithmetic questions. Finally, the Reflection \& Refinement task presents the model with a question and its CoT solution, asks it to identify potential errors, and then revise the solution into a refined CoT. In summary, SMART transfers these classic, empirically grounded cognitive frameworks to the LLM setting, providing a cognitively motivated basis for our four-dimensional evaluation.

\begin{table*}
\centering
\resizebox{\textwidth}{!}{%
\renewcommand\arraystretch{1.1}
\begin{tabular}{lccccccc} 
\toprule[1.5pt]
\multirow{2}{*}{Dataset} & \multirow{2}{*}{Final Answer} & \multicolumn{4}{c}{Cognitive Stages} & \multirow{2}{*}{Source}  & \multirow{2}{*}{Size}  \\  \cmidrule(l){3-6} 
&  & Understanding & Reasoning  & Arithmetic & R\&R &  &  \\ 
\hline
GSM8k~\cite{cobbe2021training}          & $\checkmark$ &               &              &              &              & Human                                                                                      & 8,792   \\
GSM1k~\cite{zhang2024careful}          & $\checkmark$ &               &              &              &              & GSM8K                                                                                      & 1,205   \\
MATH~\cite{hendrycks2measuring}           & $\checkmark$ &               &              & &   & 2 datasets                                                                                   & 12,500  \\
MINIF2F~\cite{zhengminif2f}      &   $\checkmark$           &               & $\checkmark$ &              &              & 4 datasets                    & 488    \\
FIMO~\cite{liu2023fimo}          & $\checkmark$             &               & $\checkmark$ &              &              & IMO                                                                                        & 149    \\
MR-GSM8k~\cite{zengmr}       &              &               &              &              & $\checkmark$ & GSM8k                                                                                      & 1,428   \\
MR-Ben~\cite{zeng2024mr}     &              &               &              &              & $\checkmark$ & 3 datasets                                & 5,975   \\
ProcessBench~\cite{zheng-etal-2025-processbench}  &              &               &              &              & $\checkmark$ & 4 datasets        & 3,400   \\ 
\hline
\textbf{SMART (ours)} & $\checkmark$ & $\checkmark$  & $\checkmark$ & $\checkmark$ & $\checkmark$ & \begin{tabular}[c]{@{}c@{}} 7 datasets \end{tabular} & 10,000  \\
\bottomrule[1.5pt]
\end{tabular}
}
\caption{Compression between our SMART and other benchmarks.}

\label{tab.smart_com_modified}
\end{table*}

\subsection{Seed Question Collection and Filtering}
The foundation of the SMART benchmark is a seed dataset comprising 2,000 problem instances. These were curated from seven widely-used mathematical problem datasets: GSM8K \cite{cobbe2021training}, SVAMP \cite{patel2021nlp}, ASDiv \cite{miao2020diverse}, AQuA \cite{ling2017program}, MAWPS \cite{koncel2016mawps}, MATH \cite{hendrycks2measuring}, and  problems from the AIME 2024 competition \cite{aime2024}. These initial 2,000 seed questions, along with their subsequently generated dimension-specific variations (four per seed question), form the complete SMART benchmark, totaling 10,000 test instances.

\begin{figure}[t]
\centering
  \includegraphics[width=\linewidth]{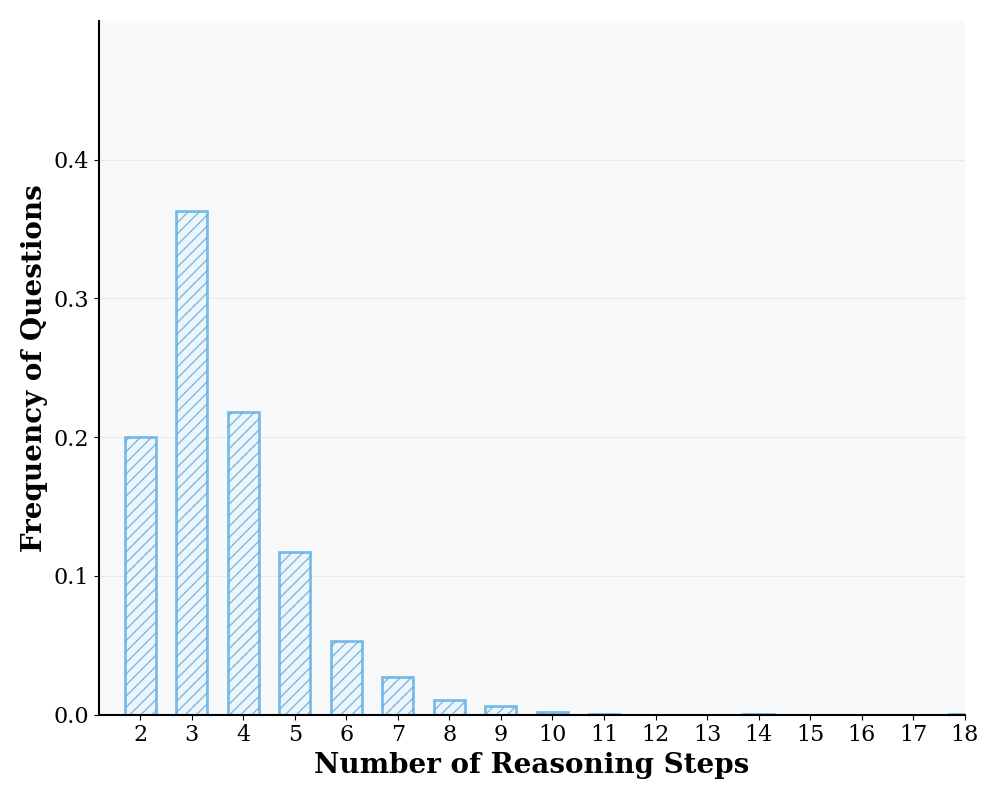}
  \caption{The reasoning step statistics of the seed question dataset.}
  \label{fig.reasonstep-stat}
\end{figure}

Several criteria were applied during the selection and processing of these seed questions. To ensure consistency in question format, problems from the AQuA dataset, originally multiple-choice, were converted into an open-ended format; the textual content of the correct option was adopted as the ground-truth for these transformed questions. Furthermore, we excluded problems whose solutions fundamentally rely on operations not readily expressible or automatically verifiable using SMT-LIB, such as calculations involving the greatest common divisor (GCD), the least common multiple (LCM), or the determination of maximum/minimum values within a set. Questions requiring multiple distinct numerical values in their answers were also omitted. To maintain a baseline level of difficulty and focus on multi-step problem-solving, we filtered out mathematical problems that could be solved in a single reasoning step. The frequency distribution of reasoning steps for the selected 2,000 seed questions is depicted in Fig.~\ref{fig.reasonstep-stat}, which illustrates that the majority of problems in the SMART benchmark involve multiple reasoning steps, with a notable concentration in the 2 to 7 step range.
\subsection{Data Annotation}
The ground truth for the majority of generated data in SMART is derived from highly reliable sources, either inherited from established benchmarks or validated through neuro-symbolic systems. Consequently, human verification efforts were strategically focused on components generated by LLMs—specifically, the structured key information for question understanding and notation-based arithmetic questions—where automatic guarantees are unavailable. Following standard practices in recent literature, we conducted manual inspections on a random subset of this data. Notably, our sampling rate of 10\% is considerably more conservative than prevalent methodologies. For instance, \cite{chen2024dr} validated Dr. Academy by manually inspecting only $\approx 1\%$ of questions; \cite{zhang2025ocr} utilized a sample of 100 Q\&A pairs per round for their OCR–RAG benchmark; and \cite{yang-etal-2025-well} inspected a random sample of 50 questions to verify unhelpful thoughts. By comparison, our 10\% re-verification protocol—under which all inspected instances were confirmed correct—provides a statistically stronger guarantee of data quality. We believe this rigorous annotation process ensures that SMART is built on a credible and sound foundation.

\subsection{Data Examples of SMART}
\begin{figure*}[t]
\centering
  \includegraphics[width=0.9\linewidth]{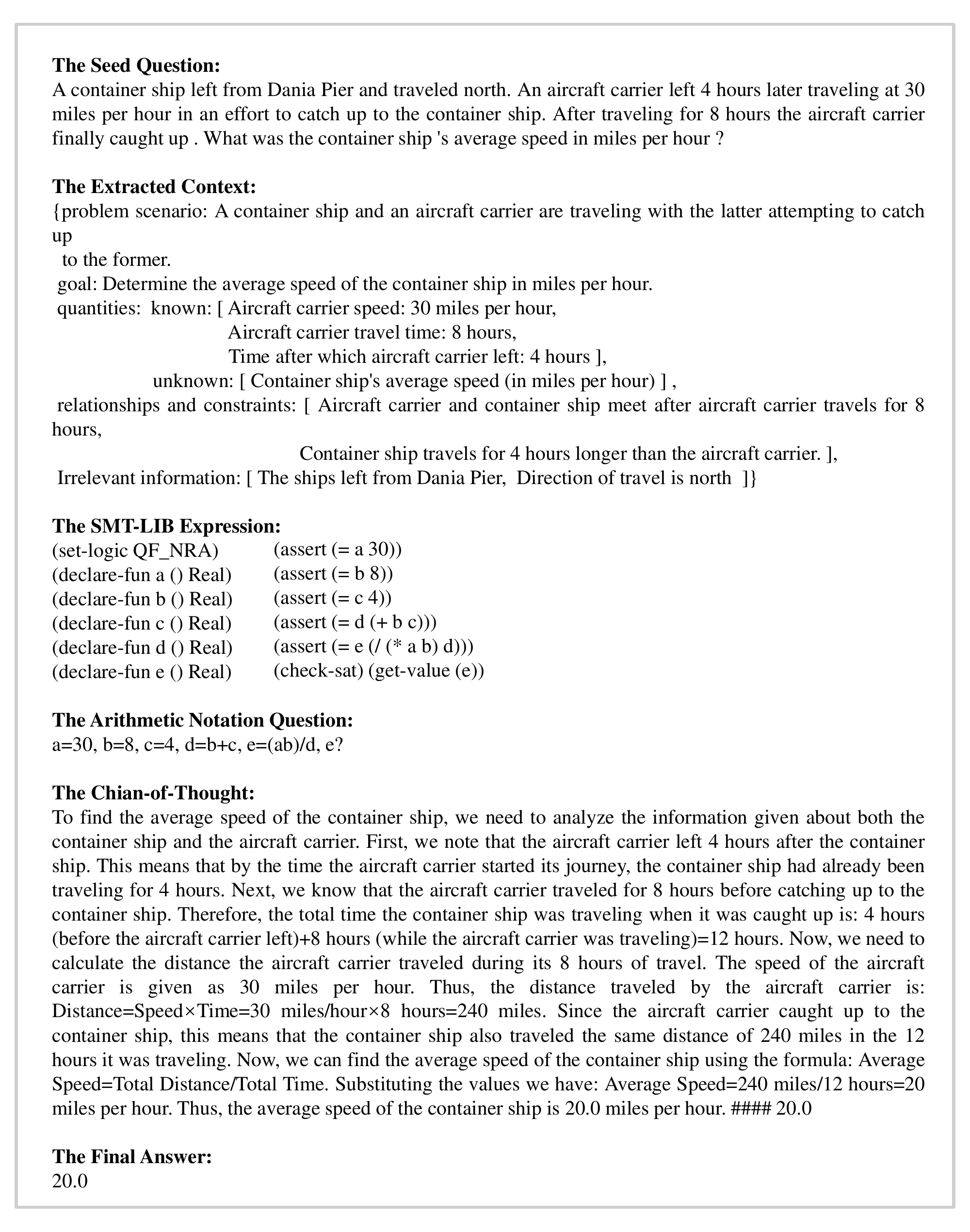}
  \caption{A data sample in the SMART benchmark.}
\label{fig.smart_data_sample}
\end{figure*}

Fig.~\ref {fig.smart_data_sample} presents a data sample of SMART, which contains the seed question, the extracted context, the SMT-LIB expression, the arithmetic notation question, the CoT, and the final answer. 

\begin{figure*}[t]
\centering
  \includegraphics[width=0.88\linewidth]{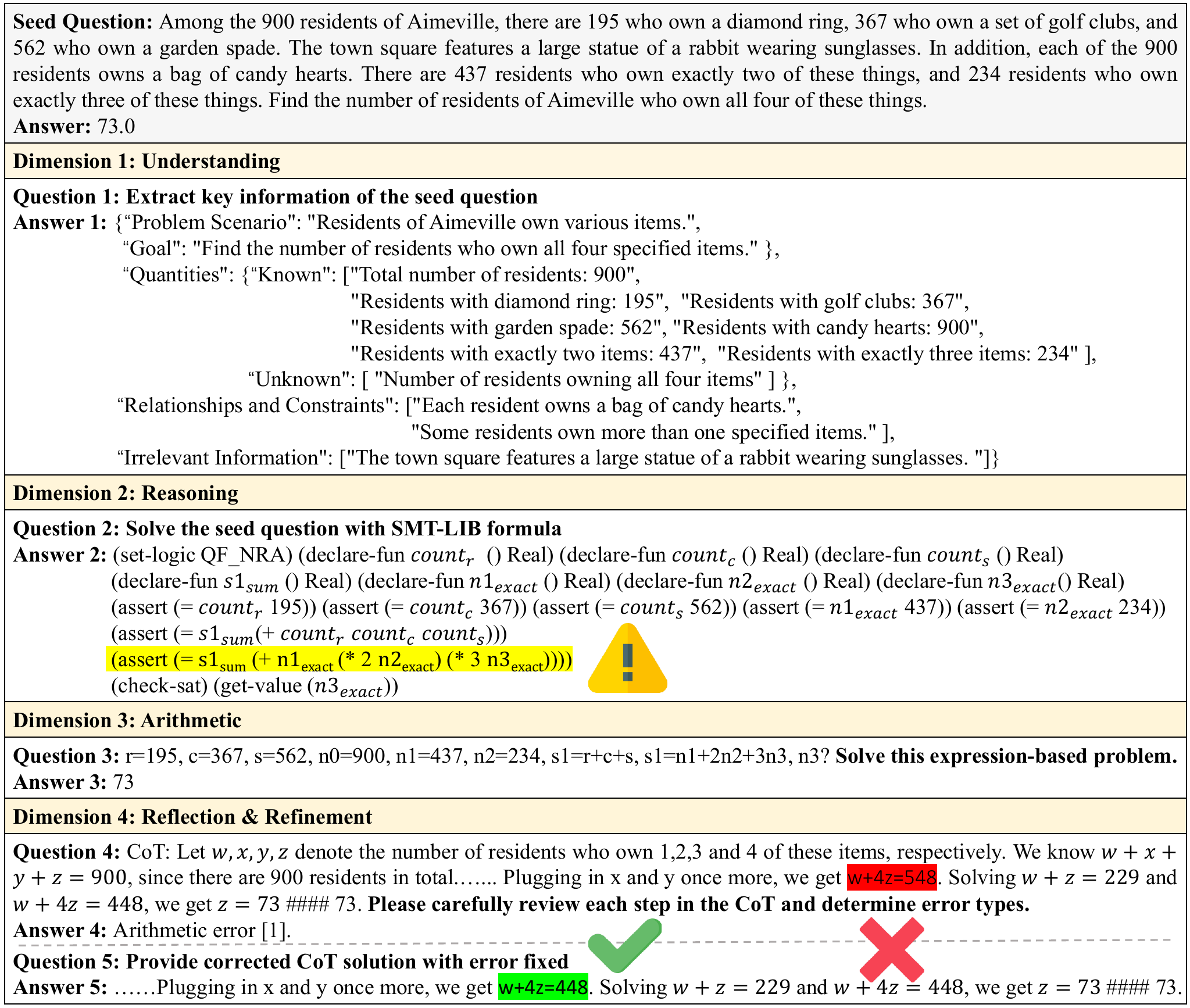}
  \caption{An overview of the SMART framework for evaluating the mathematical problem-solving process. The SMART contains four distinct dimensions. Each dimension is evaluated using dimension-specific tasks and metrics, ensuring a comprehensive assessment of the model's problem-solving capabilities. The yellow-highlighted symbolic assertion illustrates an inferred condition that is not explicitly stated in the original problem.}
  \label{fig:pipline2}
\end{figure*}

Fig.\ref{fig:pipline2} illustrates a seed question and its four-dimensional tasks. In the Understanding task, the model extracts key information from the seed question. In the Reasoning task, it solves the problem by producing an SMT-LIB formulation. In the Arithmetic task, it answers the corresponding notation-based arithmetic question. In the Reflection \& Refinement task, it first identifies error categories in the provided CoT (Reflection) and then generates a corrected CoT (Refinement). 

\subsection{Differences to Existing Benchmarks}
\label{append.dif}
In Tab. \ref{tab.smart_com_modified}, we have summarized how SMART differs from existing datasets. SMART is the first benchmark whose design is aligned with the multi-dimensional human cognitive process of mathematical problem solving. Guided by Pólya’s problem-solving theory, SMART systematically decomposes each problem along the thinking pipeline into four cognitive dimensions—Understanding, Reasoning, Arithmetic, and Reflection \& Refinement. In contrast, prior fine-grained benchmarks lack theoretical guidance and typically cover only one or two dimensions. GSM8K, GSM1k, and MATH assess LLMs solely by final-answer correctness. MINIF2F and FIMO evaluate the reasoning ability by producing formal proofs. MR-GSM8K, MR-Ben, and ProcessBench evaluate step-by-step solution verification. 
\subsection{Prompts and Rules for SMART Data Curation}
For each seed question, SMART generates distinct variants to evaluate the four targeted problem-solving dimensions. The generation and ground-truth creation for each dimensional task are described below.

\subsubsection{Understanding}
\begin{figure*}[t]
\centering
  \includegraphics[width=0.88\linewidth]{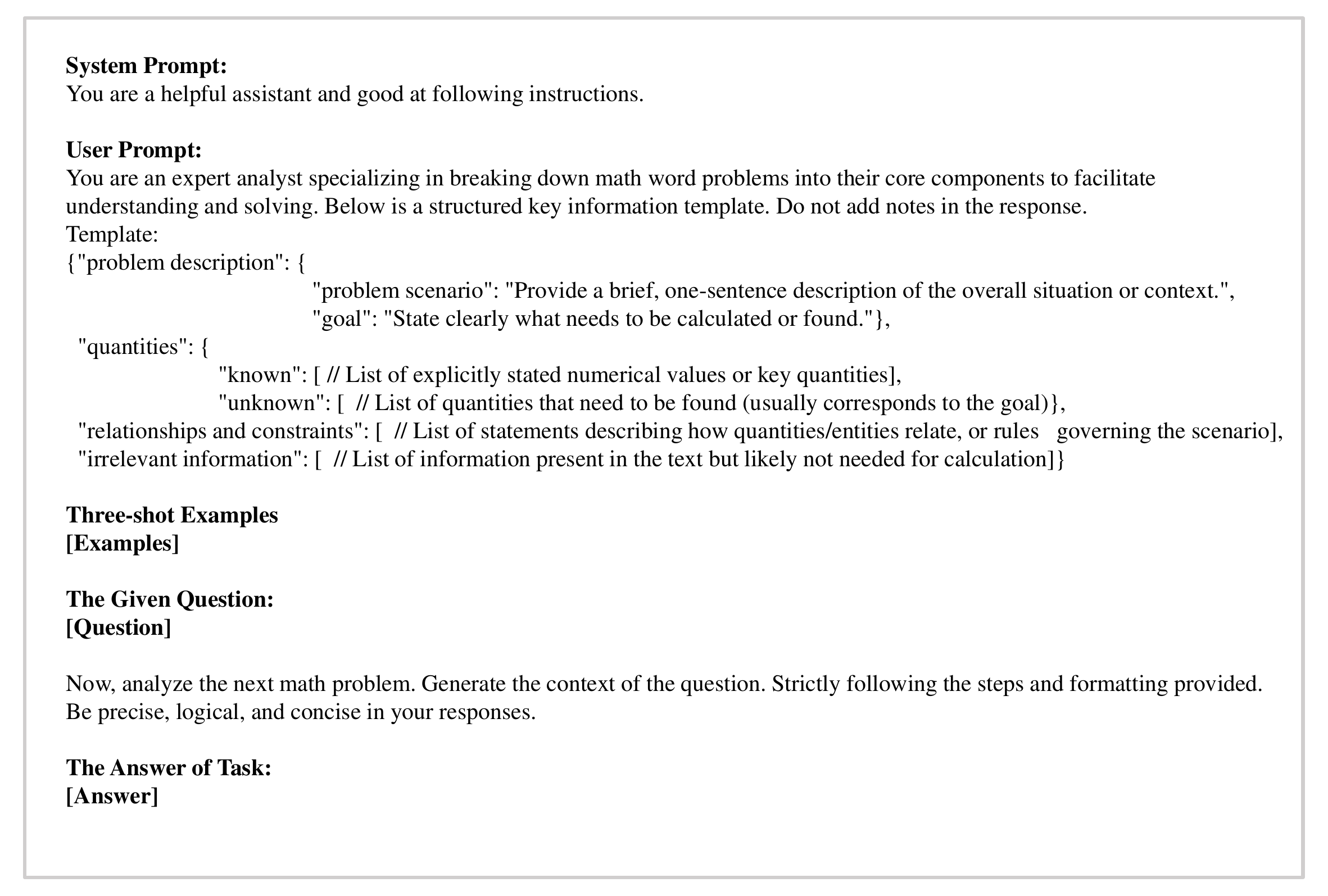}
  \caption{The prompt for LLMs to extract context from a seed question.}
  \label{fig.prompt_un}
\end{figure*}

To generate ground-truth for the understanding dimension, we utilize GPT-4.1 to perform context extraction. The extracted context is structured into the following components: Problem Scenario (describing the overall context of the problem), Goal (specifying what needs to be solved), Known Quantities (listing explicitly provided numerical values or facts), Unknown Quantities (identifying variables or values to be determined), Relationships and Constraints (detailing connections and limitations between pieces of information), and Irrelevant Information (pinpointing details not pertinent to the solution). The prompt employed to guide GPT-4.1 in extracting this contextual information for ground-truth generation is presented in Fig.~\ref{fig.prompt_un}.

\subsubsection{Arithmetic}
\begin{figure*}[t]
\centering
  \includegraphics[width=0.9\linewidth]{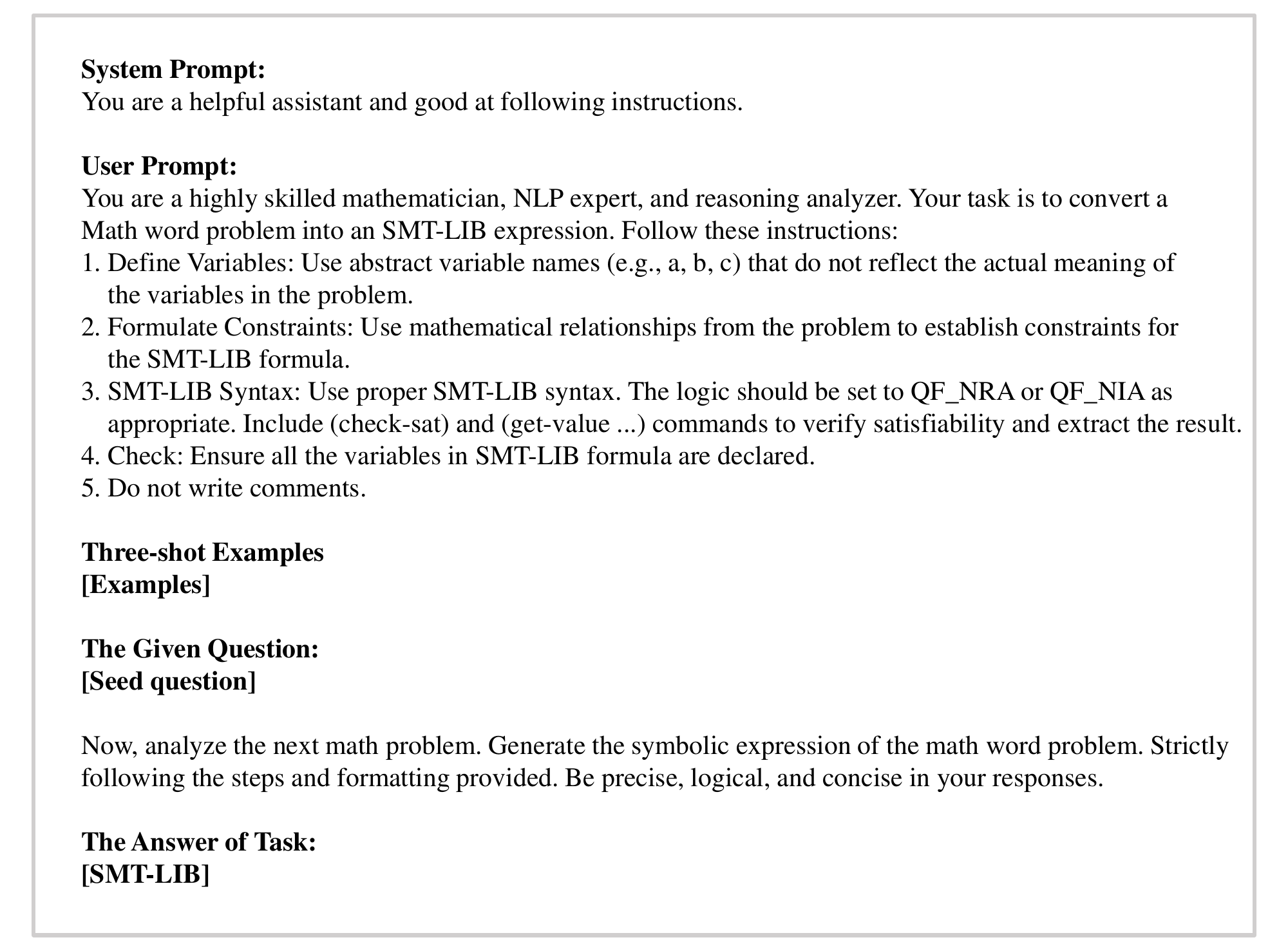}
  \caption{The prompt for LLMs to convert the seed question to a symbolic expression.}
  \label{fig.prompt_rea}
\end{figure*}

\begin{figure*}[t]
\centering
  \includegraphics[width=0.9\linewidth]{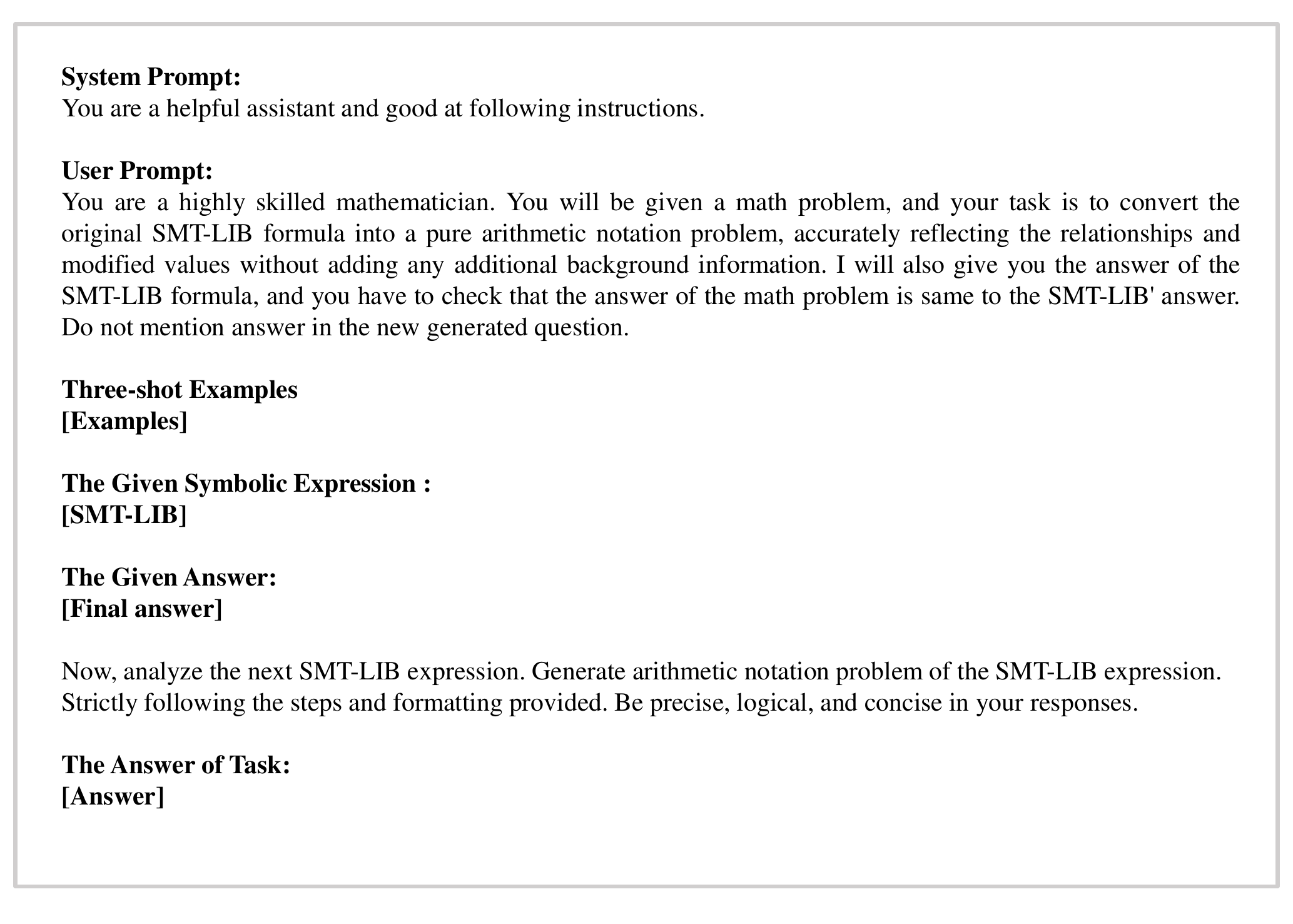}
  \caption{The prompt for LLMs to convert the SMT-LIB expression to an arithmetic notation problem.}
  \label{fig.prompt_smt2ari}
\end{figure*}

The arithmetic capability of LLMs is measured by their performance on notation-based arithmetic questions that share the same reasoning logic and final answer as the seed question. Directly converting a seed question into an arithmetic notation problem is challenging for LLMs. This difficulty arises because it requires simplifying complex natural language into structured mathematical operations while preserving the logical relationships between variables. Such a transformation is non-trivial, as the model must accurately interpret the problem's intent, handle ambiguous phrasing, and correctly map linguistic constructs to arithmetic operations. Therefore, to address this, we first generate an SMT-LIB representation of the seed question, which simplifies the reasoning logic among variables. Subsequently, we utilize GPT-4.1 to convert this SMT-LIB representation into the arithmetic notation problem, and then manually checked by human annotators. The prompt for that process is shown in Fig.~\ref {fig.prompt_rea} and Fig.~\ref{fig.prompt_smt2ari}.

\subsubsection{Reflection \& Refinement}
\begin{figure*}[t]
\centering
  \includegraphics[width=0.9\linewidth]{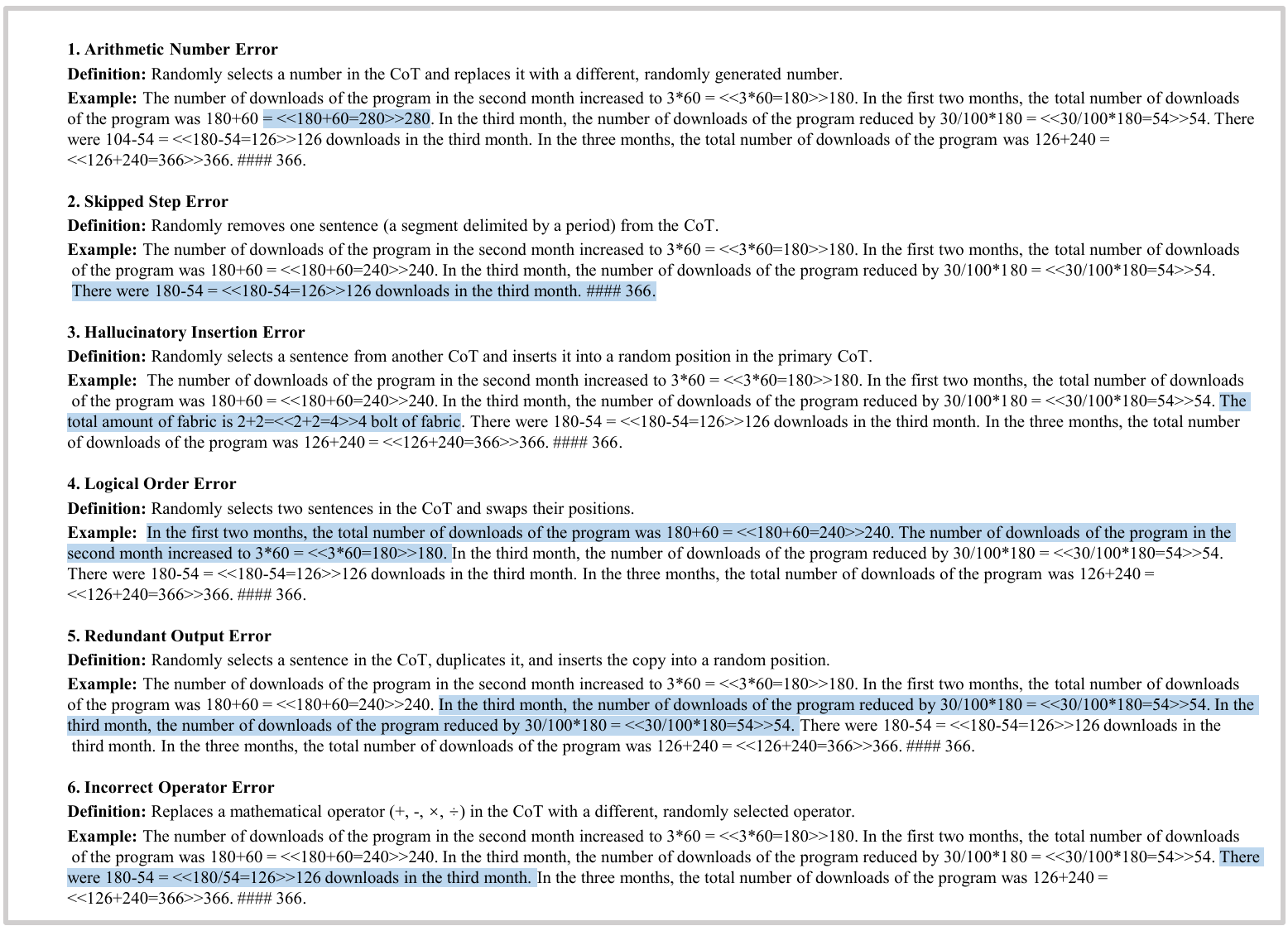}
  \caption{Example of CoT with different errors.}
  \label{fig.cotmistaks2}
\end{figure*}

For the Reflection \& Refinement dimension, we first generate CoT solutions containing deliberate errors. To create these erroneous CoTs, one of the six error types defined in Fig.~\ref{fig.cotmistaks2} (\eg arithmetic number error and skipped step) is uniformly sampled and injected into a correct CoT. 

For the Refinement task, direct verification of the LLM-corrected CoT is complex. Instead, we evaluate the refined CoT by extracting the final answer of the refined CoT. The LLM is considered to have successfully passed the Refinement task if the extracted final answer matches the ground-truth of the original seed question.

\begin{table*}[]
\resizebox{\textwidth}{!}{%
\renewcommand\arraystretch{1.2}
\begin{tabular}{l|cc|cc|cc|cc|cc}
\toprule[1.5pt]
\multirow{2}{*}{Model} &
  \multicolumn{2}{c|}{Final  Answer} &
  \multicolumn{2}{c|}{Understanding} &
  \multicolumn{2}{c|}{Reasoning} &
  \multicolumn{2}{c|}{Arithmetic} &
  \multicolumn{2}{c}{R\&R} \\ \cline{2-11} 
                        & Zero-shot & Three-shot & Zero-shot & Three-shot & Zero-shot & Three-shot & Zero-shot & Three-shot & Zero-shot & Three-shot \\ \hline
GPT-4.1                 & 43.25     & 59.34      & 90.25     & 95.41      & 59.52     & 73.29      & 38.65     & 45.71      & 37.05     & 59.90      \\
o4-mini                 & 92.25     & 93.75      & 93.58     & 95.22      & 85.63     & 90.72      & 97.12     & 98.25      & 51.20     & 55.05      \\
\begin{tabular}[c]{@{}l@{}}Gemini2.5\\ -Flash-Preview\end{tabular} & 90.65     & 93.33      & 91.12     & 94.25      & 81.47     & 89.72      & 95.41     & 98.18      & 30.74     & 35.85      \\ 
\bottomrule[1.5pt]
\end{tabular}}
\caption{Zero-shot and three-shot prompt engineer strategy comparison in SMART.}
\label{append.zero}
\end{table*}

\begin{table}[]
\resizebox{\columnwidth}{!}{%
\renewcommand\arraystretch{1.1}
\begin{tabular}{l|cc}
\toprule[1.5pt]
Method                & CoT to SMT-LIB & Question to SMT-LIB \\ \hline
GPT-4.1               & 92.58          & 73.29               \\
O4-mini               & 95.68          & 90.72               \\
DeepSeek-R1           & 97.54          & 93.61               \\
Claude3.7-Sonnet      & 96.25          & 74.18               \\
Gemini2.5-Pro & 97.24          & 91.45               \\
\bottomrule[1.5pt]
\end{tabular}}
\caption{Accuracy for transferring CoT and Seed question to SMT-LIB formula}
\label{tab.abr}
\end{table}

\begin{figure*}[t]
\centering
  \includegraphics[width=0.9\linewidth]{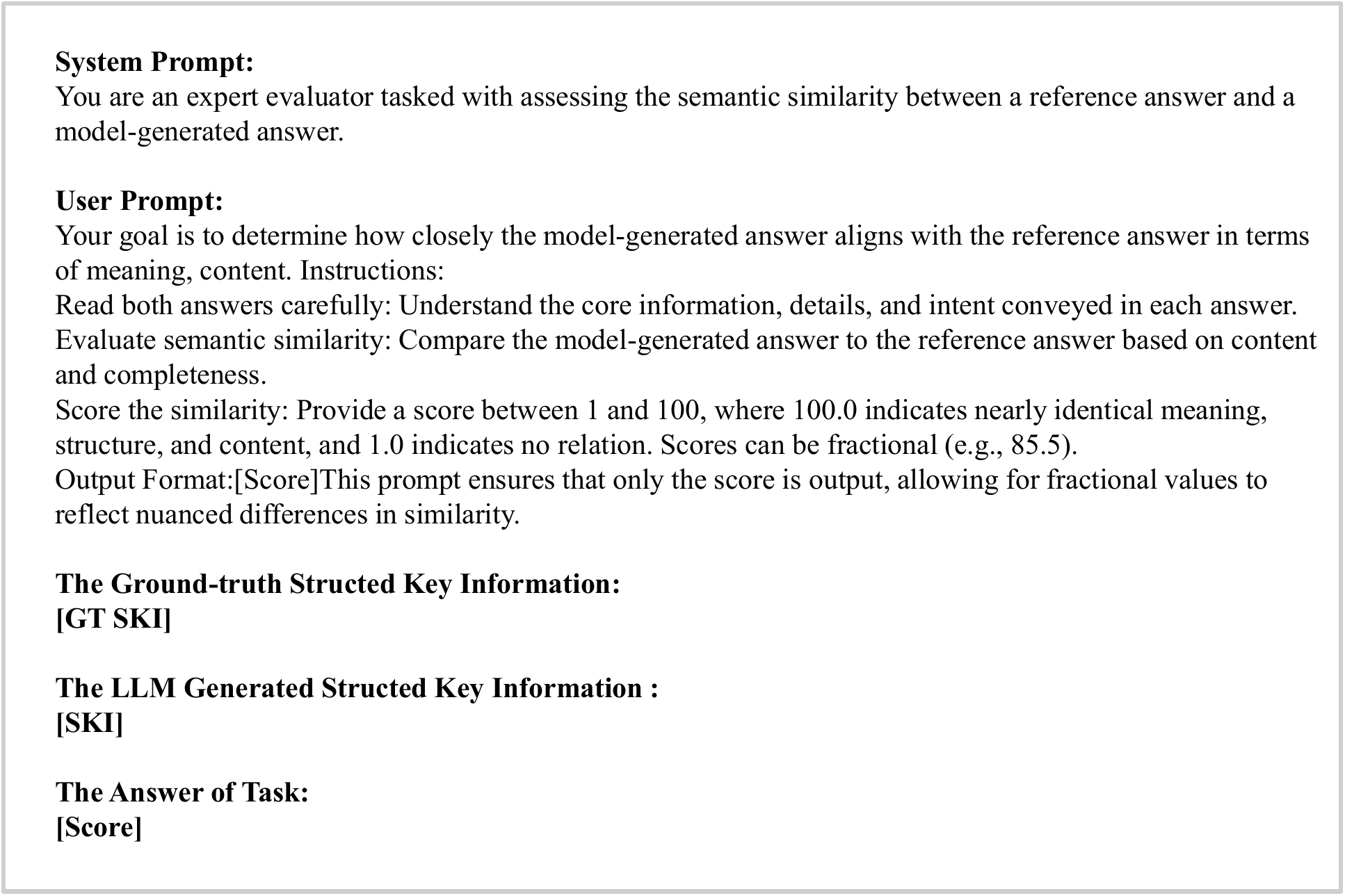}
  \caption{The prompt for LLM-as-a-Judge for evaluating the Understanding task.}
  \label{fig.prompt_llmasjudge}
\end{figure*}

\begin{figure*}[t]
\centering
  \includegraphics[width=0.9\linewidth]{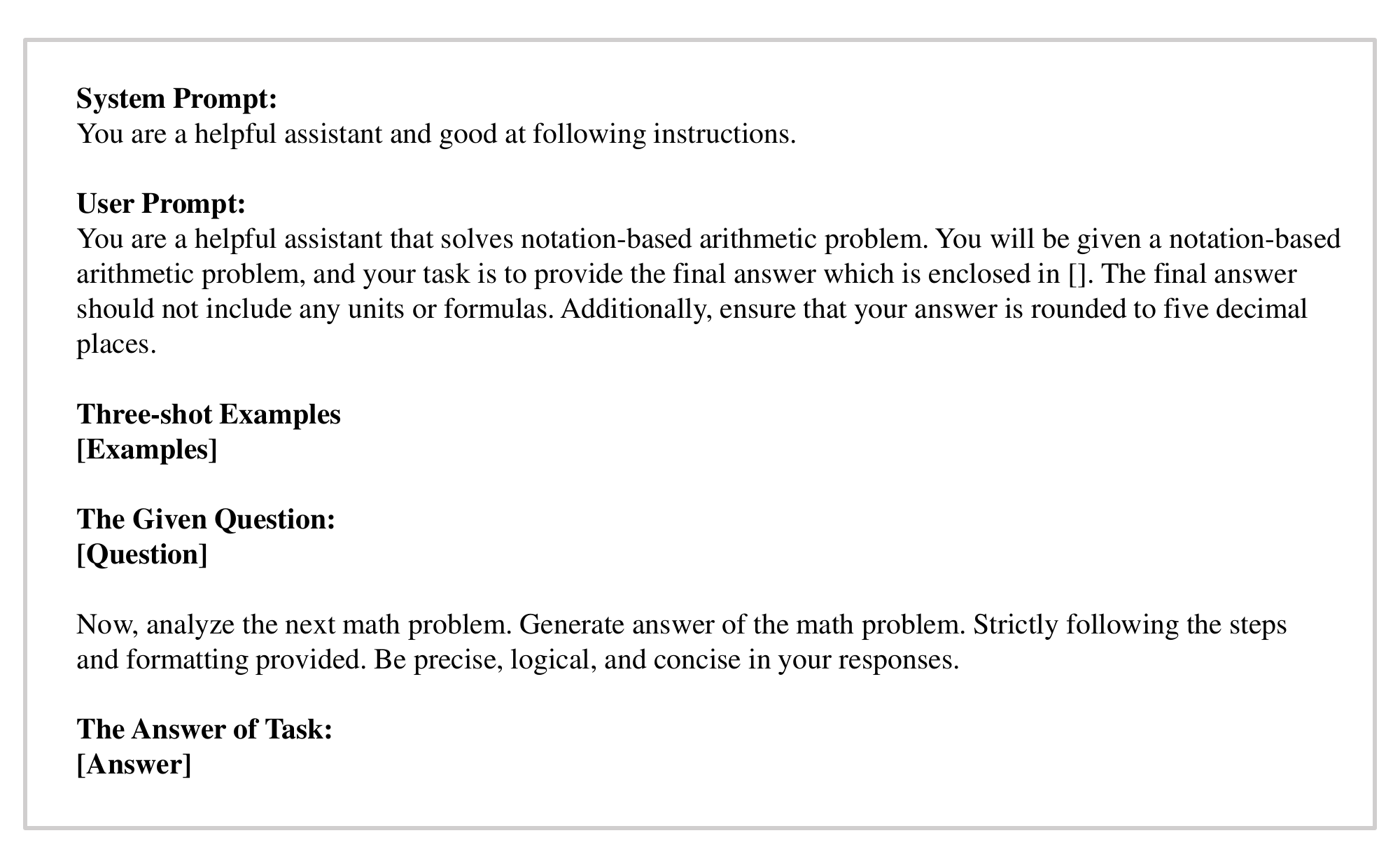}
  \caption{The prompt for LLMs to solve the arithmetic notation problem.}
\label{fig.prompt_solve}
\end{figure*}

\begin{figure*}[t]
\centering
  \includegraphics[width=0.9\linewidth]{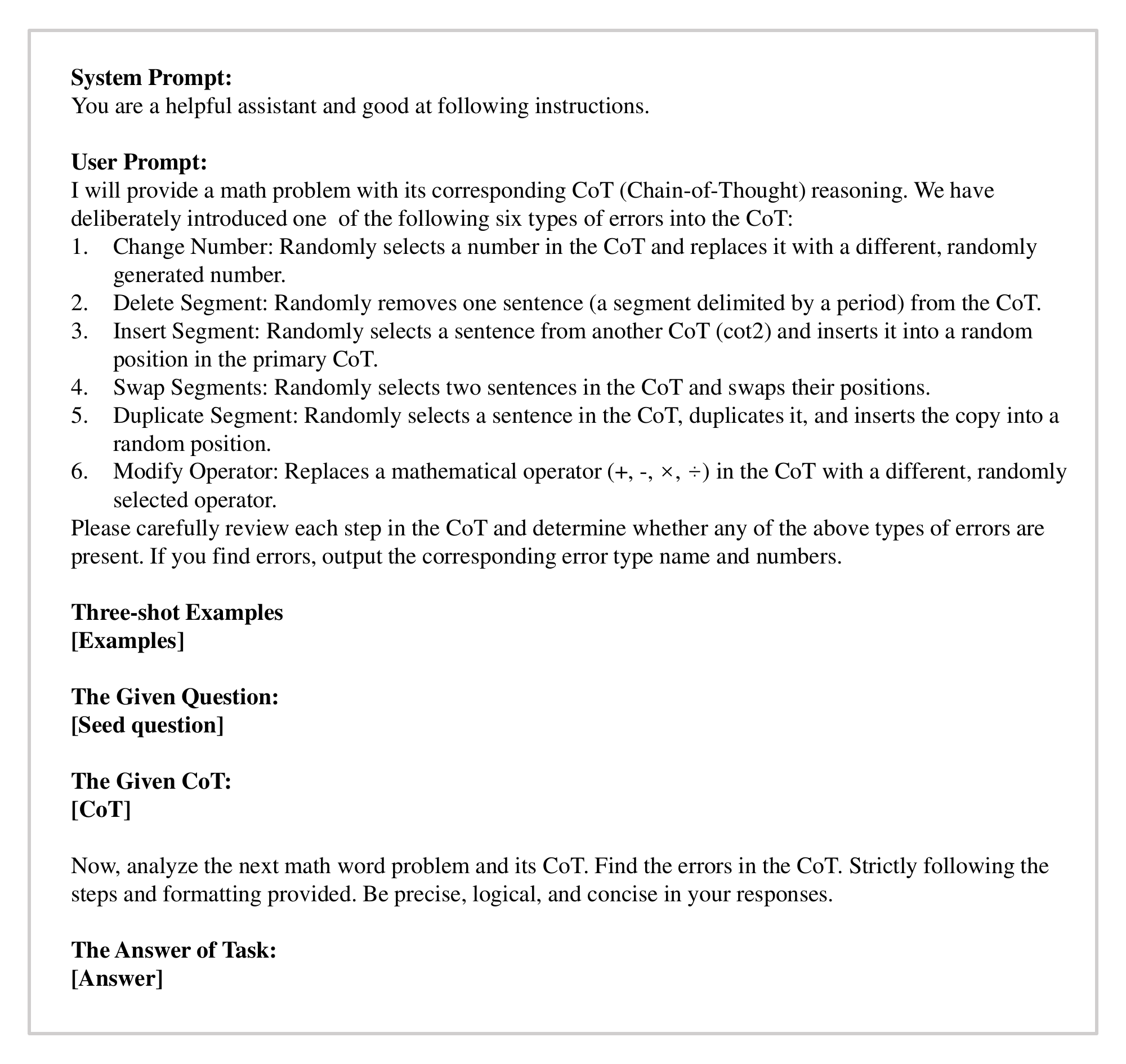}
  \caption{The prompt for LLMs to detect mistakes in the CoT.}
  \label{fig.prompt_det}
\end{figure*}
\begin{figure*}[t]
\centering
  \includegraphics[width=\linewidth]{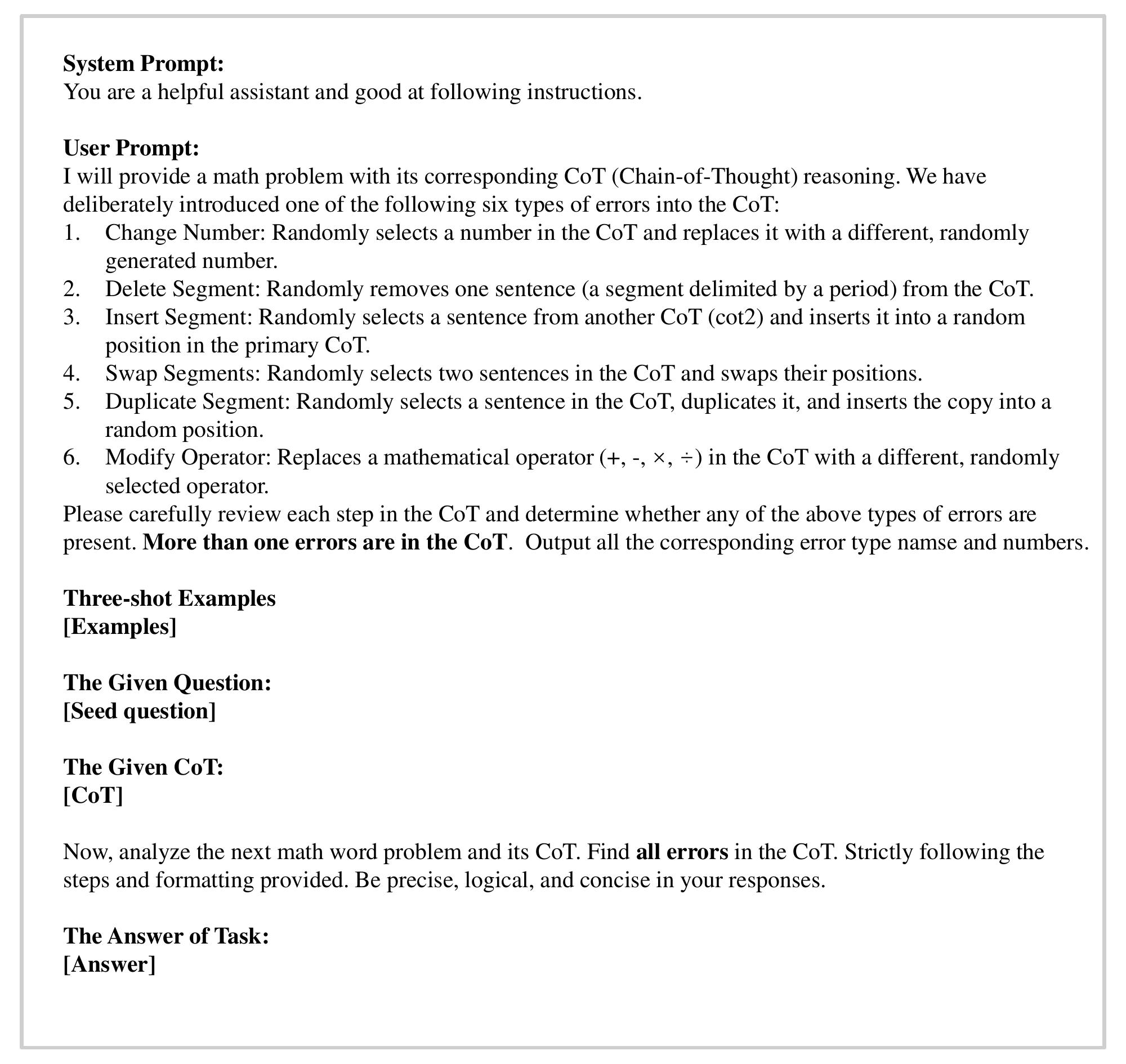}
  \caption{The prompt for LLMs to detect more than one mistake in the CoT.}
  \label{fig.prompt_det2}
\end{figure*}
\begin{figure*}[t]
\centering
  \includegraphics[width=\linewidth]{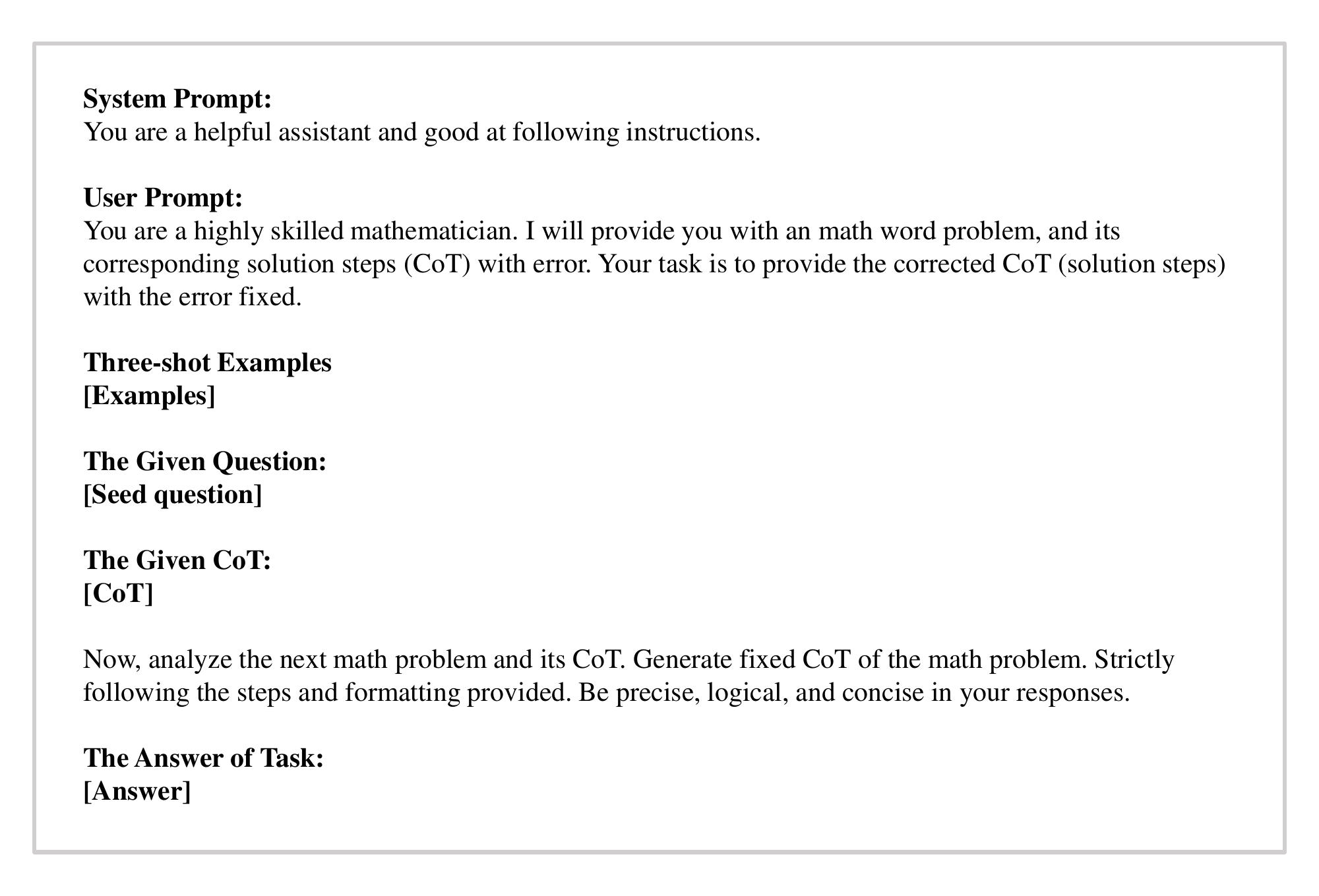}
  \caption{The prompt for LLMs to correct the mistakes in the CoT.}
  \label{fig.prompt_corrcot}
\end{figure*}

\subsection{Experiment Setting}
We evaluate 22 recent open-source and closed-source LLMs using our SMART evaluation framework. The open-source models include Phi4 \cite{abdin2024phi}, Gemma3 \cite{team2025gemma}, GLM4 \cite{glm2024chatglm}, Qwen2.5 \cite{qwen2.5}, Qwen3 \cite{yang2025qwen3}, Llama3 \cite{llama3modelcard}, Llama4 \cite{llama4}, and Mistral \cite{mistral-small}. The closed-source models assessed are GPT-4o \cite{achiam2023gpt}, GPT-4.1 \cite{OpenAIgpt4.1}, o4-mini, o3, GPT-5 \cite{OpenAIo3-o4mini}, DeepSeek-V3 \cite{liu2024deepseek}, DeepSeek-R1 \cite{guo2025deepseek}, Claude3.5 \cite{claude35}, Claude3.7 \cite{claude37}, and Gemini2.5 \cite{gemin25}. 

All experiments were conducted on a Linux server equipped with two NVIDIA H800 GPUs (80GB). The GPUs were used for deploying and performing inference on open-source models. The Python version used was 3.9.20, and the version of the Transformers package was 4.46.0.

\subsection{Prompts for Evaluation  in SMART}
\label{append.smart_eval}
\subsubsection{Understanding}
We evaluate the generated structured key information by comparing it with the ground-truth structured key information and then use LLMs as the judge model to give the similarity score. The evaluation prompt is presented in the   Fig.~\ref{fig.prompt_llmasjudge}.

\subsubsection{Reasoning}
For the reasoning dimension, we introduce a symbolic formalization task to evaluate the symbolic reasoning capability of LLMs. The question for this task is the seed question, and LLMs are asked to generate the SMT-LIB expression of the question, without solving the problem. The prompt for this task is shown in Fig.~\ref{fig.prompt_rea}. Subsequently, the Z3 Solver is used to compute the results of the generated SMT-LIB expression. Finally, we compare the results of SMT-LIM expression to the ground-truth of seed questions.

\subsubsection{Arithmetic}
For the arithmetic dimension, we introduce a numeric calculation task to evaluate the arithmetic capability of LLMs. The question for this task is a notation-based arithmetic problem, and LLMs are asked to solve it using the prompt shown in Fig.~\ref{fig.prompt_solve}. Then, we compare the results of the notation-based questions to the ground-truth of seed questions.

\subsubsection{Reflection \& Refinement}
For the reflection \& refinement dimension, we propose an error correction task that requires LLMs to detect mistakes in the chain-of-thought (CoT) of seed questions, correct these mistakes, and generate a new answer for the seed question.  The first step involves detecting errors in the CoT, given the question and CoT, with the answer being the specific name of the introduced error type. The evaluation prompt is shown in Fig. \ref{fig.prompt_det} and prompt for more errors in Fig. \ref{fig.prompt_det2}. If LLMs fail to detect all mistakes, they do not need to attend the following refinement task. The second step is to fix errors in CoT and generate a refined CoT with the prompt shown in Fig.\ref{fig.prompt_corrcot}. The final step is to extract the new final answer of the refined CoT with rule-matching. If LLMs successfully detect all mistakes and generate the correct final answer based on the corrected CoT, we consider the model to have passed the error correction task.

\subsection{The Self-Refine Prompting}
\begin{figure*}[t]
\centering
  \includegraphics[width=\linewidth]{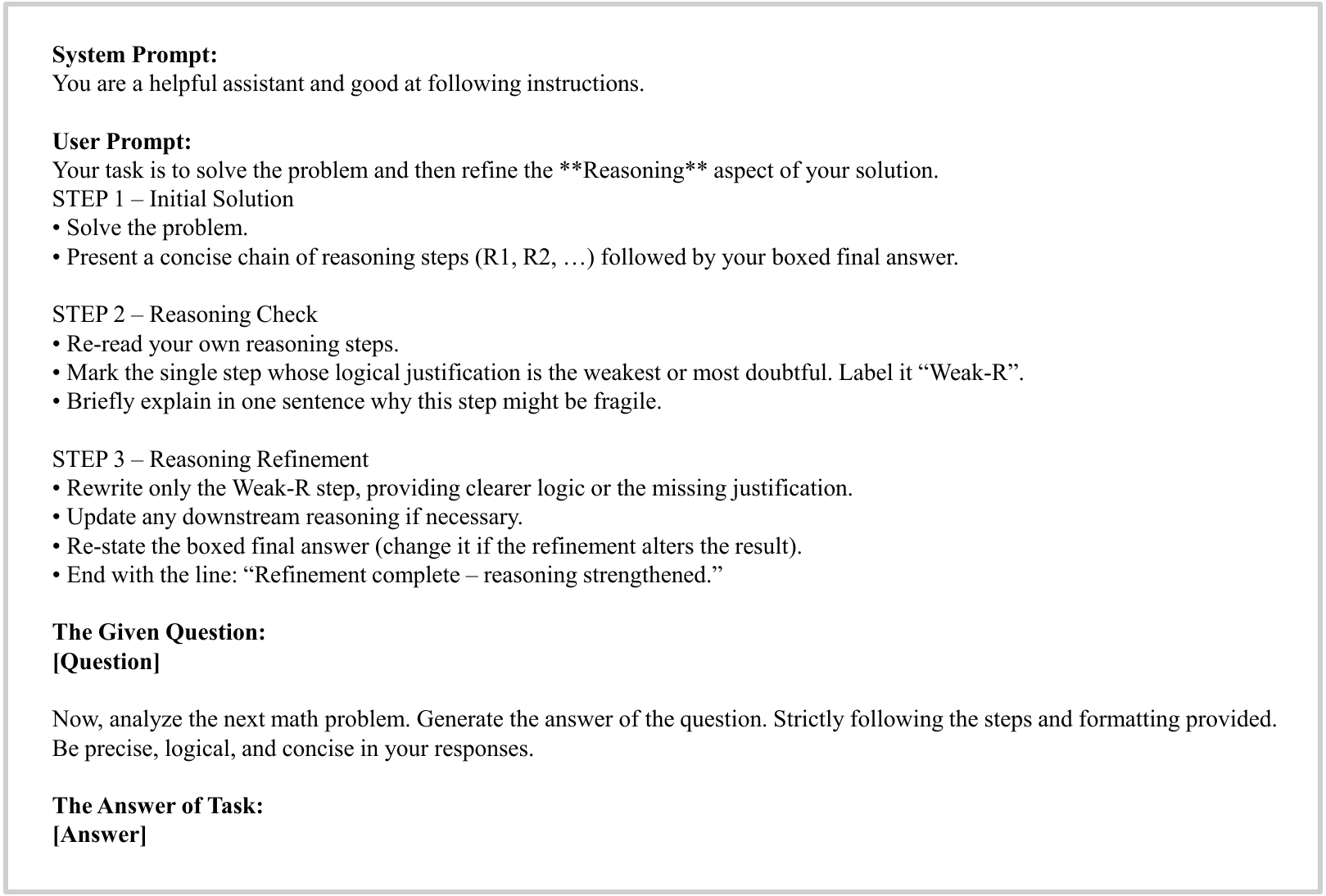}
  \caption{The prompt for LLMs to self-refine the Reasoning dimension.}
  \label{fig.selfpromptrea}
\end{figure*}

\begin{figure*}[t]
\centering
  \includegraphics[width=\linewidth]{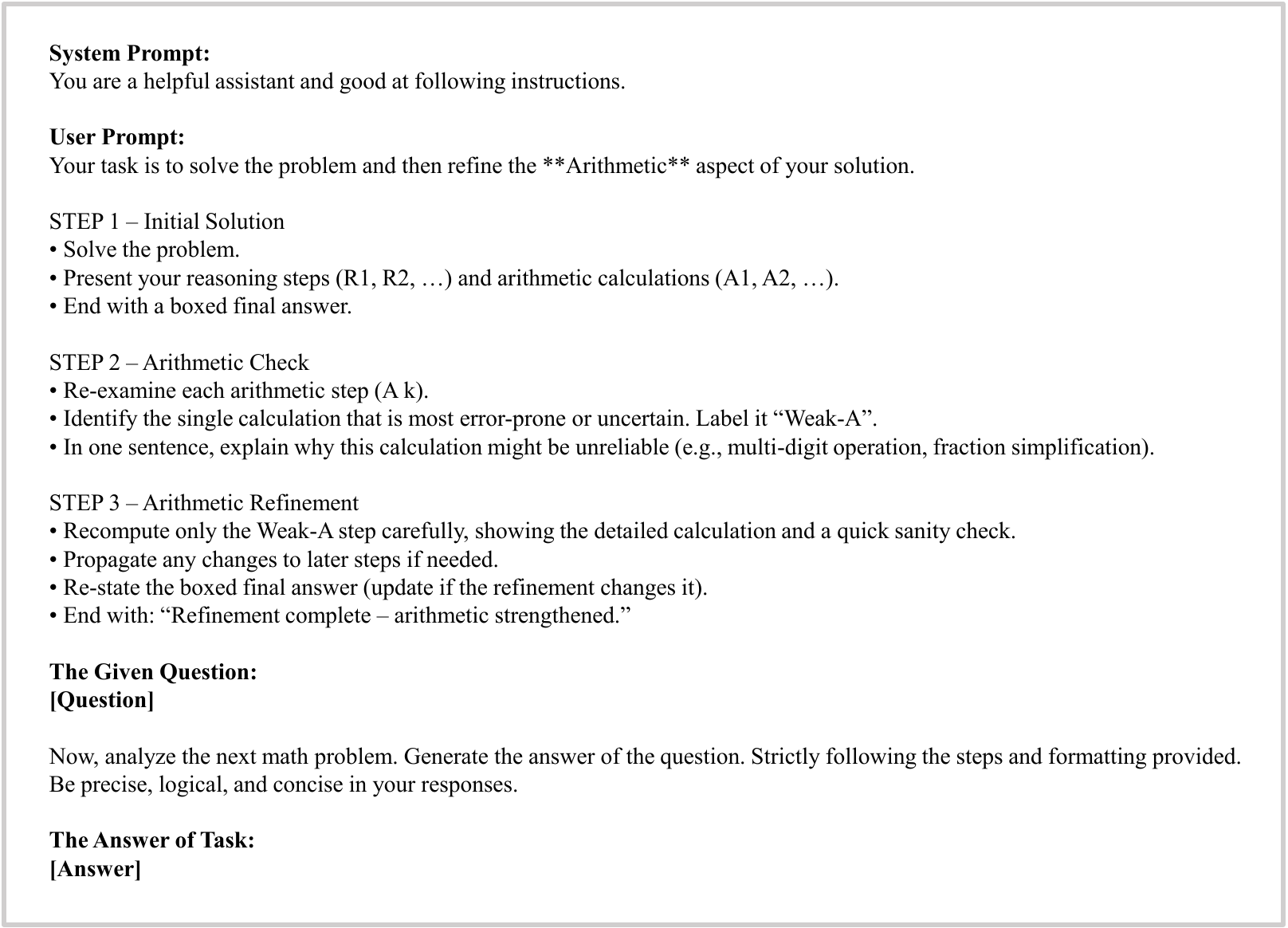}
  \caption{The prompt for LLMs to self-refine the Arithmetic dimension.}
  \label{fig.selfpromptari}
\end{figure*}

We apply self-refinement prompting specifically to the weakest step identified in the problem-solving process to improve the mathematical capability of LLMs. The prompts used for self-refinement on the reasoning and arithmetic dimensions are shown in Fig.\ref{fig.selfpromptrea} and Fig.\ref{fig.selfpromptari}.

\subsection{Difficulty setting for Dimension-specific Task}
\label{append.smart-dyn1}
To generate dimension-specific questions with varying difficulty, we employ the following strategies:

For the understanding dimension, difficulty is controlled by progressively introducing irrelevant sentences, sourced from other problems, as 'noise' within the seed question's text. The number of such noise sentences dictates the complexity of the context extraction task. The ground-truth for these modified questions is updated by incorporating these noise sentences into the 'Irrelevant Information' category of the context extraction template.

For the reasoning dimension, question complexity is defined by the number of distinct 
mathematical operations (\eg $+,-,\times,\div,mod$) required to formulate the solution. Problems are then categorized into multiple difficulty levels based on this operational count.

In the arithmetic dimension, complexity is varied by altering the number of digits in the numerical values involved (\eg changing '12' to a five-digit number like '34.823'), rather than solely their magnitude, as precision with more digits presents a distinct challenge. The ground-truth for these modified arithmetic problems is obtained through our quality control method.

In the reflection \& refinement dimension, difficulty is modulated by the number and types of mistakes deliberately injected into the Chain-of-Thought solutions. We randomly introduce a varying number of distinct error types (from the categories listed in Fig.~\ref{fig.cotmistaks2}) into the CoT to create different levels of challenge for the error detection and correction tasks.

Each seed question undergoes a two-stage transformation process. In the first stage, it is decomposed into four distinct, dimension-specific tasks that enable fine-grained evaluation of individual capabilities. In the second stage, these tasks are further rewritten using validated augmentation strategies to generate diverse variants that test the robustness and adaptability of LLMs. This hierarchical, two-phase generation framework enhances reliability and scalability by enabling comprehensive and granular evaluation while mitigating risks of overfitting and data contamination.
\subsection{Examples for questions with different difficulty settings}
\label{append.diff}
\begin{figure*}[t]
\centering
  \includegraphics[width=\linewidth]{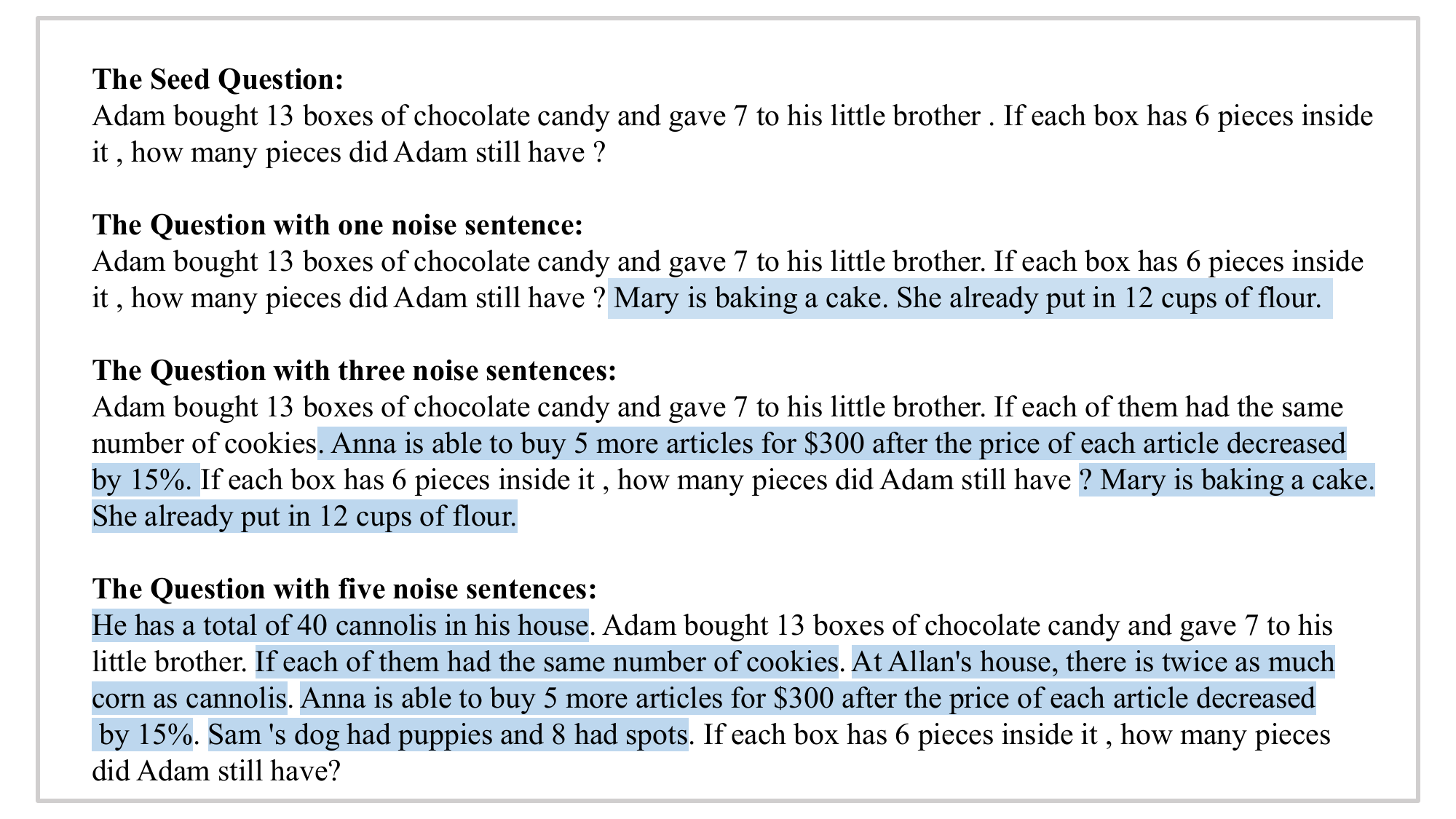}
  \caption{Example of questions with a different number of noise sentences.}
  \label{fig.noise}
\end{figure*}

Fig.\ref{fig.noise} shows examples of different difficulty settings for the understanding dimension evaluation. The sentences with a red background in the image represent irrelevant noise sentences, and the more noise sentences there are, the harder the task of extracting the effective context becomes.

\begin{figure*}[t]
\centering
  \includegraphics[width=\linewidth]{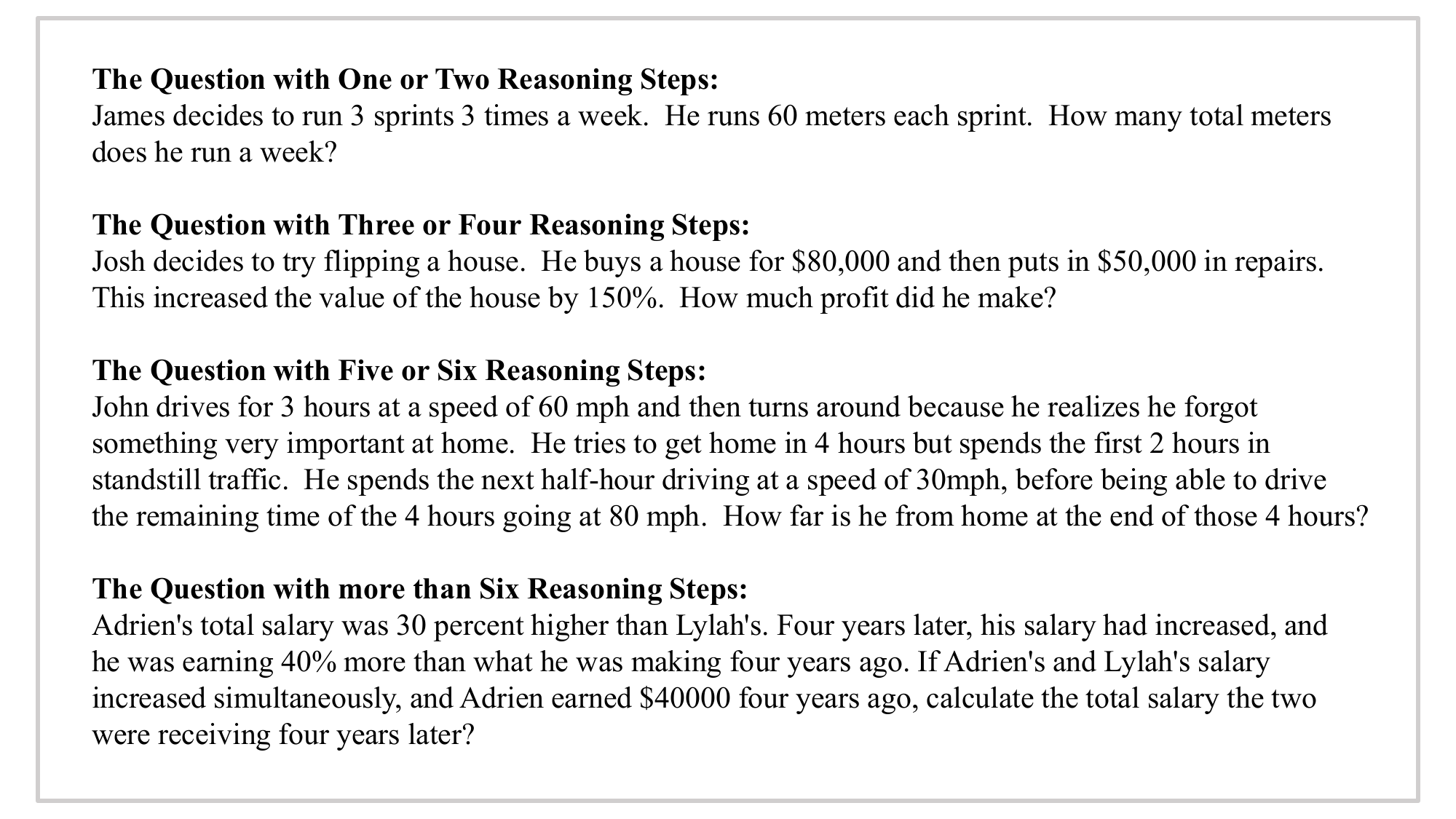}
  \caption{Example of questions with different reasoning steps.}
  \label{fig.reasonsteps}
\end{figure*}

Fig.\ref{fig.reasonsteps} shows examples of questions with different reasoning steps, which indicate different reasoning difficulties.

\begin{figure*}[t]
\centering
  \includegraphics[width=\linewidth]{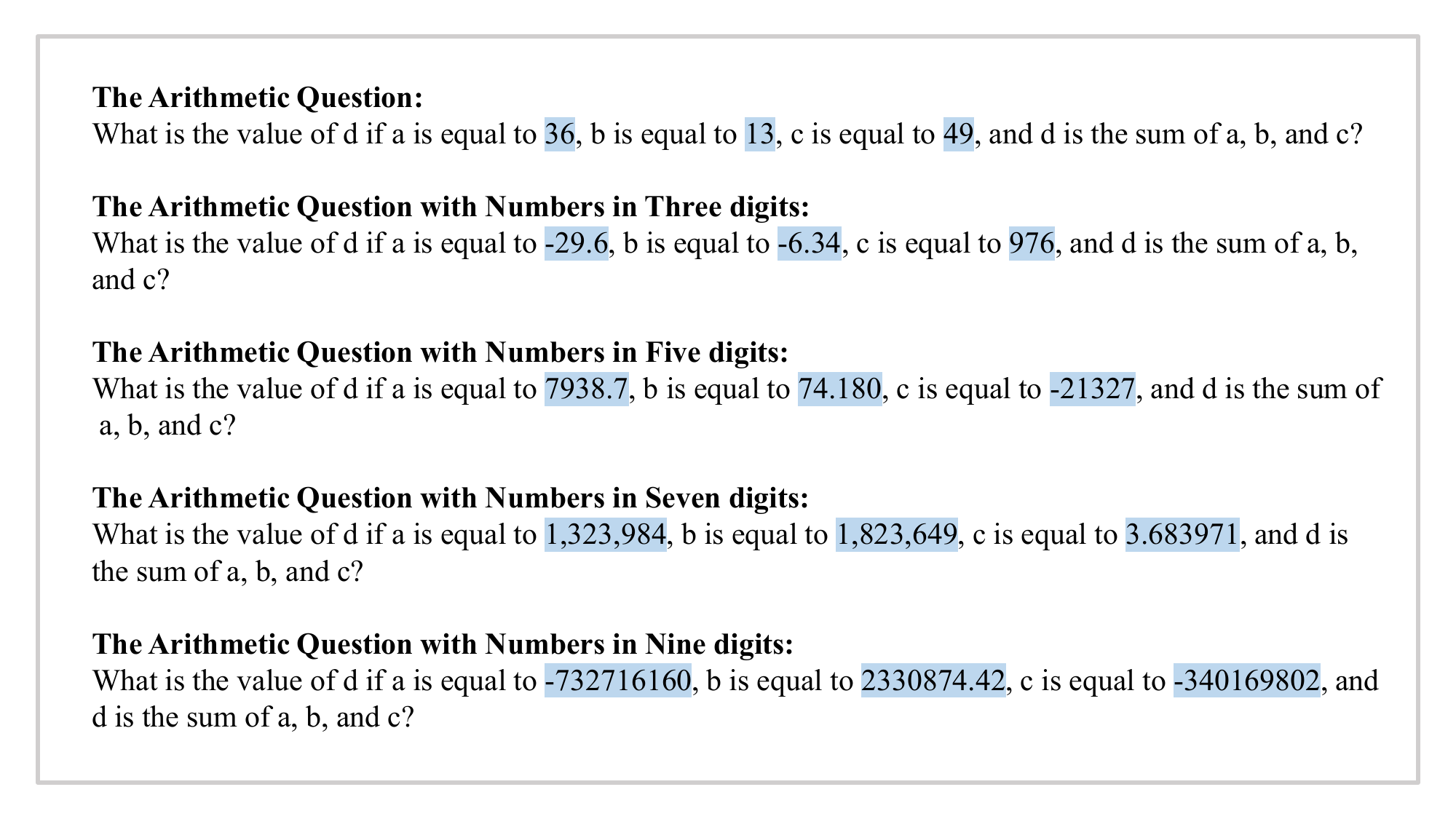}
  \caption{Example of arithmetic questions with numbers in different digits.}
  \label{fig.digits}
\end{figure*}

Fig.\ref{fig.digits} presents arithmetic questions with numbers in different digits. Numbers with more digits are more difficult for the arithmetic evaluation task.

Fig.\ref{fig.cotmistaks2} presents CoT with different types of errors. CoT with more mistakes is more difficult for the reflection and refinement evaluation task.

\begin{figure*}[t]
\centering
  \includegraphics[width=\linewidth]{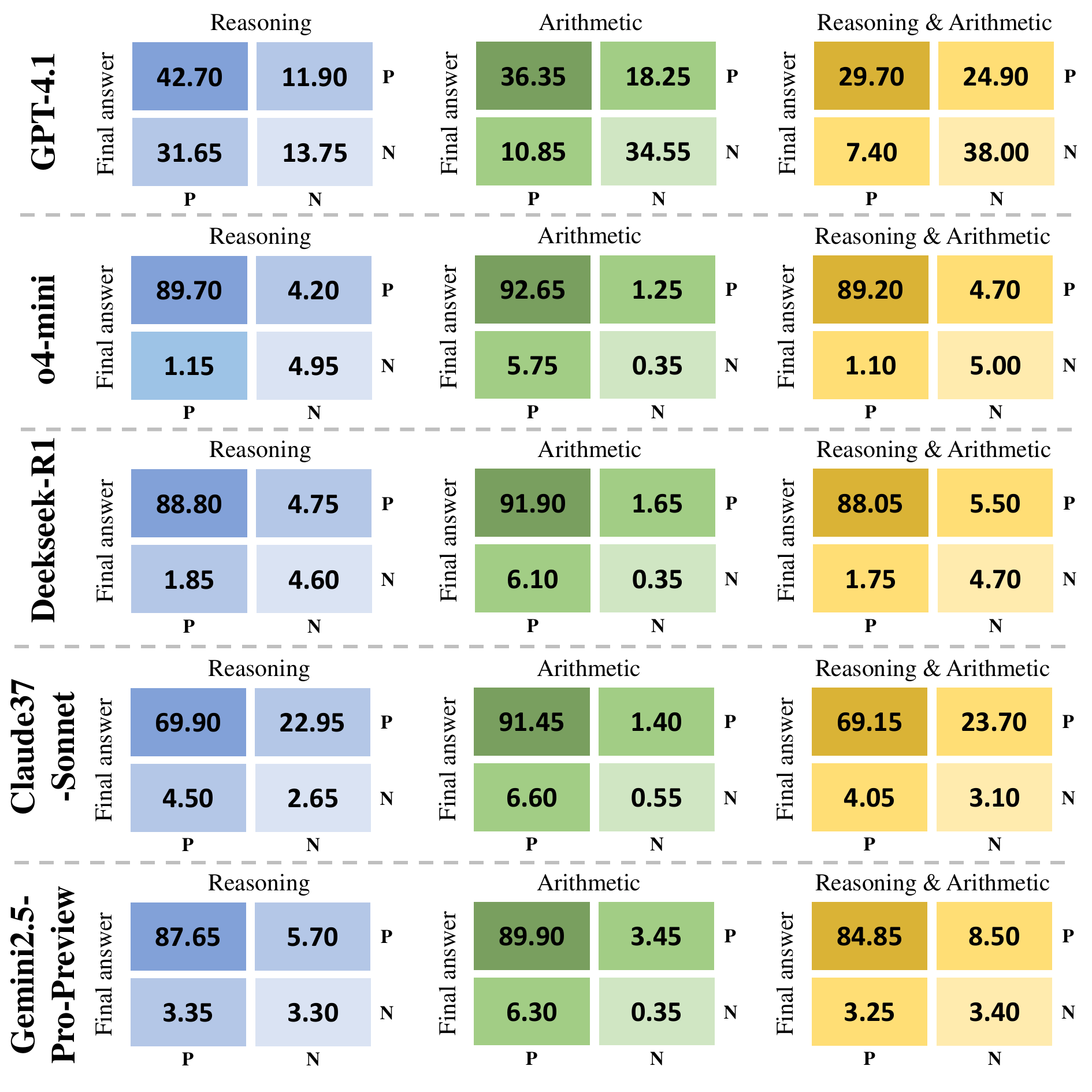}
  \caption{The confusion matrix of the final answer and other dimensions. P means Positive, and N means Negative.}
  \label{fig:conf}
\end{figure*}

\subsection{Zero-shot vs. three-shot}
We conducted additional comparative experiments evaluating model performance with and without three-shot examples across all stages. The results are presented in the Tab.\ref{append.zero}. As shown in the table, removing the three-shot examples leads to a significant performance drop across all models for dimensions. This confirms the substantial positive impact of including a few-shot example for these more complex reasoning stages. Our comparative results show that adding a few-shot example provides an improvement in all dimension scores across.

\subsection{Ablation Study for Reasoning Task}
LLMs demonstrate strong capabilities in translating natural language into formal language, and the formal translation process itself is not the primary cause of poor performance on the Reasoning task. To verify this, we conducted an ablation study ('CoT-to-SMT-LIB') where models converted correct natural language CoT into SMT-LIB formulas in Tab.\ref{tab.abr}. Accuracy was verified via Z3 execution against ground-truth answers. All evaluated models achieved over 90\% accuracy, indicating that the main challenge lies in generating correct reasoning paths—not in the formal translation. This supports our claim that the Reasoning dimension effectively captures a model’s core mathematical Reasoning capability.

\subsection{Analysis of Potential Circularity Bias in Evaluation} We address the concern regarding potential circularity—specifically, the risk of self-preference bias—arising from the use of GPT-4.1 for both ground-truth generation and evaluation in the Understanding task. To rigorously quantify this effect, we conducted a sensitivity analysis using an independent set of judges.

We introduced a new judge ensemble consisting of DeepSeek-V3 and Claude 3.5 Sonnet. This ensemble is distinct from the primary setup (GPT-4.1 + DeepSeek-V3) used in the main paper. We re-evaluated the top-performing models and compared the scores averaged from the new independent ensemble against the original scores.

As shown in Table \ref{tab:circularity_check}, the scoring remains highly consistent across different judge configurations. The absolute difference between the scores yielded by the independent ensemble (DeepSeek-V3 + Claude) and the original ensemble (GPT-4.1 + DeepSeek-V3) is at most 0.08.

This negligible deviation demonstrates that the Understanding scores reported in SMART are robust to the choice of judge models. The inclusion of GPT-4.1 in the evaluation loop does not introduce statistically significant circularity bias, ensuring the validity of our leaderboard rankings.
\begin{table*}[]
\centering
\begin{tabular}{l|ccccc}
\toprule[1.5pt]
Model             & DeepSeek-V3 & Claude 3.5 & Avg (DS, Claude) & Avg (GPT-4.1, DS) & $\Delta$    \\ \hline
GPT-4.1           & 95.19       & 95.25      & 95.22            & 95.3              & 0.08 \\
o4-mini           & 95.28       & 95.14      & 95.21            & 95.25             & 0.04 \\
GPT-5             & 95.17       & 95.32      & 95.25            & 95.18             & 0.07 \\
DeepSeek-V3       & 95.12       & 95.15      & 95.14            & 95.12             & 0.02 \\
Claude 3.5 Sonnet & 94.81       & 94.65      & 94.73            & 94.75             & 0.02 \\
Gemini 2.5-Flash  & 95.04       & 94.45      & 94.75            & 94.69             & 0.06 \\
Qwen2.5-72B       & 95.58       & 95.89      & 95.74            & 95.86             & 0.08 \\ \bottomrule[1.5pt]
\end{tabular}
\caption{The performance of the Understanding task with different LLMs as the judge models.}
\label{tab:circularity_check}
\end{table*}
\subsection{Analysis of Dimensional Independence}
To validate the empirical difference of tasks in SMART, we conduct an empirical analysis to determine whether the four evaluation dimensions (Understanding, Reasoning, Arithmetic, R\&R) provide distinct signals or collapse into a single capability metric. We analyze the performance of 22 models using both quantitative correlation matrices and qualitative ranking discrepancies.

We compute the Spearman correlation coefficient ($\rho$) and associated $p$-values between all pairs of dimensions. As shown in Table \ref{tab.spear}, the results indicate that the dimensions capture relatively independent capabilities:\begin{itemize}\item \textbf{Understanding is Distinct:} The Understanding dimension exhibits a near-zero correlation with Arithmetic ($\rho = 0.06, p = 0.79$) and only a moderate, statistically non-significant correlation with Reasoning ($\rho = 0.36, p = 0.10$). This suggests that the ability to parse and structurally comprehend a problem is functionally distinct from the ability to execute symbolic operations.\item \textbf{Coupling of Execution Capabilities:} The Reasoning, Arithmetic, and R\&R dimensions show moderate correlations. This is expected, as successful reasoning often relies on correct arithmetic execution, and reflection (R\&R) requires re-evaluating both reasoning and calculation. However, the correlations are far from perfect, implying that they still measure distinguishable aspects of the problem-solving process.\end{itemize}

The independence of these dimensions is further evidenced by substantial shifts in model rankings across different tasks. Discrepancies in rankings allow for a fine-grained diagnosis of model-specific bottlenecks:\begin{itemize}\item Case Study 1: Qwen2.5-72B. This model ranks 1st in Understanding but drops to 22nd in Arithmetic. This highlights a "semantic-strong but computation-weak" profile, where the model excels at interpreting questions but fails at basic execution.\item Case Study 2: DeepSeek-V3. Conversely, this model ranks 4th in Arithmetic and 2nd in R\&R, yet places 12th in Reasoning. This suggests robust computational and self-correction mechanisms, with mathematical reasoning planning being the primary bottleneck.\end{itemize}These findings confirm that SMART’s multi-dimensional framework provides a holistic and granular view of model capabilities, avoiding the oversimplification of a single aggregate score.
\begin{table*}[h]
    \centering

    \label{tab:spearman_corr}
    \resizebox{1\linewidth}{!}{
    \begin{tabular}{l c c c c}
        \toprule[1.5pt]
        Dimension & Understanding & Reasoning & Arithmetic & R\&R \\
        \midrule
        Understanding & 1.00 & 0.361 \small{(0.0986)} & 0.059 \small{(0.7949)} & 0.611 \small{(\textbf{0.0025})} \\
        Reasoning    & -    & 1.00 & 0.669 \small{(\textbf{0.0003})} & 0.605 \small{(\textbf{0.0002})} \\
        Arithmetic    & -    & -    & 1.00 & 0.664 \small{(\textbf{0.0008})} \\
        R\&R          & -    & -    & -    & 1.00 \\
        \bottomrule[1.5pt]
    \end{tabular}
    }
    \caption{Spearman correlation coefficients ($\rho$) between the four evaluation dimensions across 22 models. $P$-values are shown in parentheses. \textbf{Bold} indicates statistical significance ($p < 0.05$).}
    \label{tab.spear}
\end{table*}
\subsection{Is the Final Answer Accuracy  Reliable for Measuring Mathematical Capability? }

Given the impressive performance of LLMs and the potential for data contamination, there is a concern that models might solve problems correctly without possessing genuine underlying mathematical capability. To investigate this, we conduct experiments to compute confusion matrices comparing final answer correctness with performance on our  SMART dimensions, as illustrated in Fig.~\ref{fig:conf}. We posit that true mathematical capability is more accurately reflected by instances where LLMs correctly solve not only the original seed question but also simultaneously succeed in the corresponding reasoning and arithmetic dimension tasks.

Across all confusion matrices, the False Negative (FN) values are consistently non-zero. This indicates that LLMs can sometimes arrive at correct final answers through heuristic shortcuts or other opaque mechanisms, even when their intermediate reasoning or calculation processes are flawed. For example, GPT-4.1 exhibits a notable FN rate of 11.90\% in the final answer and arithmetic confusion matrix. Similarly, Claude3.7-Sonnet shows an FN rate of 22.95\% in the final answer and reasoning confusion matrix. These FN cases represent instances where final answer accuracy overestimates the model's grasp of the intermediate steps. 

Conversely, False Positive (FP) scores denote cases where a model successfully completes an intermediate task but ultimately yields an incorrect final answer. Except for GPT-4.1, most evaluated LLMs exhibit relatively low FP rates across various confusion matrices, indicating that accurate intermediate reasoning generally correlates with correct final outputs.

True Positive (TP) scores capture instances where a model not only produces the correct final answer but also performs intermediate reasoning and arithmetic correctly. We regard this TP metric as a more reliable indicator of genuine mathematical problem-solving capability. For high-performing models such as o4-mini, DeepSeek-R1, and Gemini2.5-Pro-Preview, TP scores in the reasoning \& arithmetic confusion matrix closely match their ACC@Fi values. In contrast, GPT-4.1 and Claude3.7-Sonnet exhibit significantly lower TP scores relative to their ACC@Fi, suggesting that their final answer accuracy may overestimate their true reasoning capabilities.

\end{document}